\documentclass[10pt,twocolumn,letterpaper]{article}

\usepackage[pagenumbers]{wacv} 

\usepackage{graphicx}
\usepackage[accsupp]{axessibility}
\usepackage{amsmath}
\usepackage{amssymb}
\usepackage{booktabs}
\usepackage{placeins}
\usepackage{tabularx}

\usepackage{adjustbox}
\usepackage{graphicx}
\usepackage{lscape}
\usepackage[table]{xcolor}

\usepackage[pagebackref,breaklinks,colorlinks]{hyperref}

\usepackage[capitalize]{cleveref}
\crefname{section}{Sec.}{Secs.}
\Crefname{section}{Section}{Sections}
\Crefname{table}{Table}{Tables}
\crefname{table}{Tab.}{Tabs.}

\def\wacvPaperID{679} 
\def\confName{WACV}
\def\confYear{2025}

\begin{document}

\title{Few-shot Structure-Informed Machinery Part Segmentation with\\
 Foundation Models and Graph Neural Networks}

\author{Michael Schwingshackl \qquad Fabio F. Oberweger \qquad Markus Murschitz\\
AIT Austrian Institute of Technology\\Center for Vision, Automation \& Control\\
{\tt\small \{michael.schwingshackl, fabio.oberweger, markus.murschitz\}@ait.ac.at}
}

\maketitle

\begin{abstract}
This paper proposes a novel approach to few-shot semantic segmentation for machinery with multiple parts that exhibit spatial and hierarchical relationships. Our method integrates the foundation models CLIPSeg and Segment Anything Model (SAM) with the interest point detector SuperPoint and a graph convolutional network (GCN) to accurately segment machinery parts. By providing 1 to 25 annotated samples, our model, evaluated on a purely synthetic dataset depicting a truck-mounted loading crane, achieves effective segmentation across various levels of detail. Training times are kept under five minutes on consumer GPUs. The model demonstrates robust generalization to real data, achieving a qualitative synthetic-to-real generalization with a $J\&F$ score of 92.2 on real data using 10 synthetic support samples. When benchmarked on the DAVIS 2017 dataset, it achieves a $J\&F$ score of 71.5 in semi-supervised video segmentation with three support samples. This method's fast training times and effective generalization to real data make it a valuable tool for autonomous systems interacting with machinery and infrastructure, and illustrate the potential of combined and orchestrated foundation models for few-shot segmentation tasks.
\end{abstract}

\section{Introduction}
\label{sec:intro}
\begin{figure}[t!]
  \centering
    \includegraphics[width=0.9\linewidth]{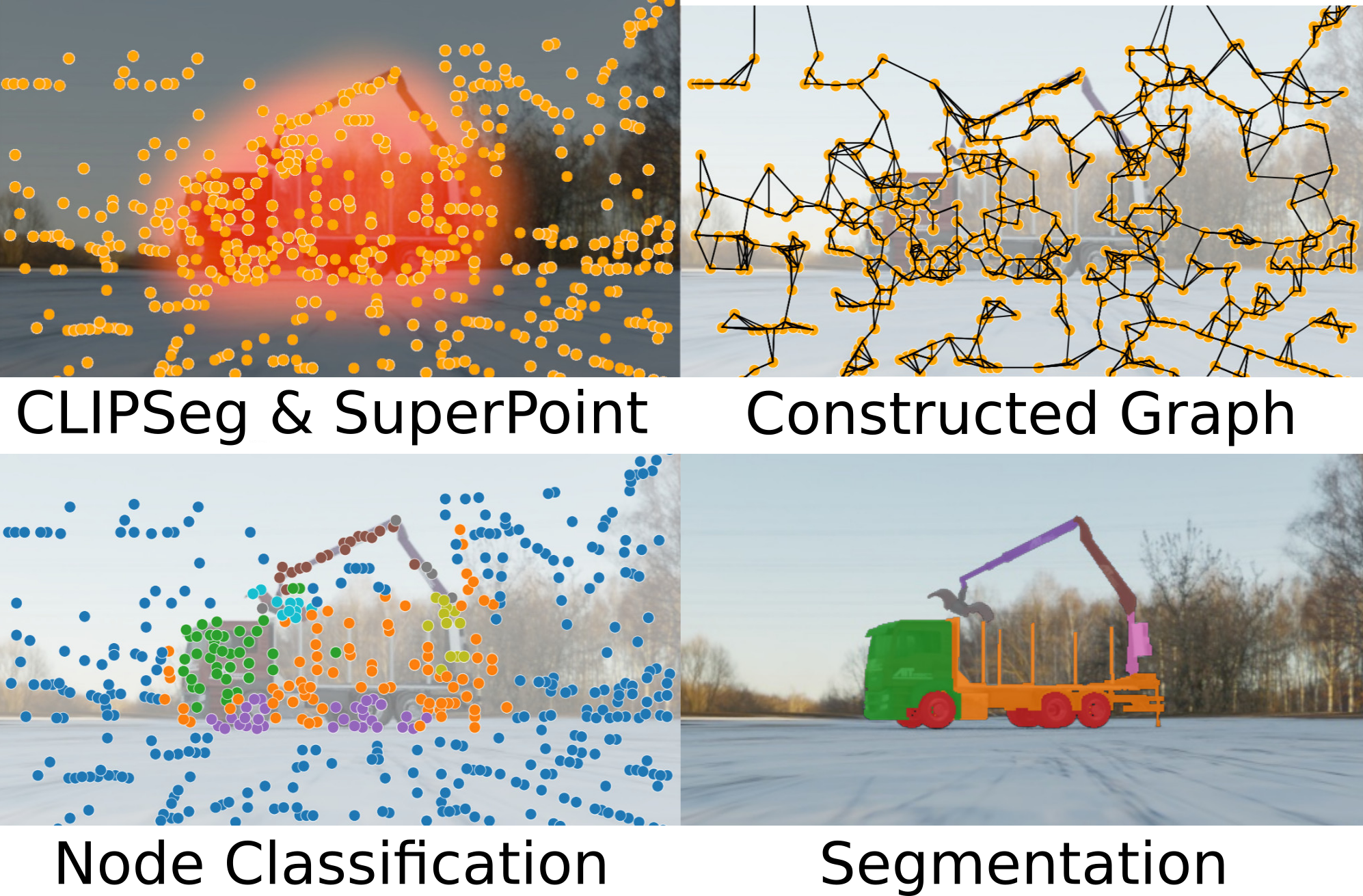}
    \caption{Intermediate steps of our pipeline.}
    \label{fig:PipelineSteps}
\end{figure}

\begin{figure*}[t]
  \centering
    \includegraphics[width=0.98\textwidth]{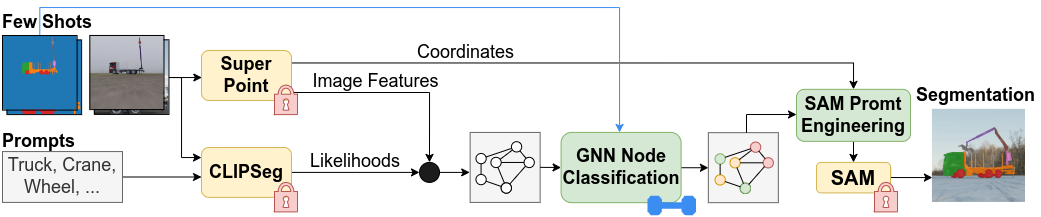}
    \caption{System architecture, with all frozen foundation models (yellow) and the novel modules (green). Only the GNN is trained.}
    \label{fig:Architecture}
    
  \hfill
\end{figure*}
To effectively manage autonomous operations, it is often required to interact with human-made functional structures composed of multiple parts such, as doors, levers, or machines in general. To perform this interaction, the perception part of an autonomous system has to identify the different parts of the structure and label them according to a corresponding concept of function. For example, to interact with a door, we need to understand which part is the handle before opening it. Also, identifying the position of individual parts of such structures can provide significant insight into the current state and potentially even into future actions of the observed structures. For instance, the orientation of a car's front wheel can indicate its intended direction of travel even during a standstill. Any human driver takes this into account, and so should automated machines, which are to operate safely.
Moreover, a modern autonomous system learns during operation and must adapt quickly to new functional structures. This is why our approach has to work with only a few labeled examples (shots) and minimal training time (a few minutes) on consumer hardware. 
To do so, we heavily rely on foundation models that recently revolutionized the field of computer vision. We use CLIPSeg \cite{luddecke2022image}, SuperPoint\cite{detone2018superpoint}, and Segment Anything (SAM)\cite{kirillov2023segment} and combine them with a small few-shot trained Graph Neural Network (GNN) to gain a robust and adaptable system. See \Cref{fig:PipelineSteps} for intermediate results.
We first demonstrate this system for structure-informed composite object part detection trained on synthetic images of a truck-mounted loading crane to prove its capability for complex composite functional structures (on synthetic and real-world data). Then, we prove its adaptability with evaluations on the DAVIS 2017 dataset, where we can produce state-of-the-art results even if the dataset is not very well suited for the problem at hand.

The application driving the truck-mounted loading crane scenario is collaborative loading with an automated ground vehicle and a manually operated crane. However, robustly finding structured compositions of objects in image data has many other applications, such as self-introspection of machines, infrastructure interaction (doors, loading areas, levers, and more), behavioral cloning, and physical model identification and parameter tuning. 

An overview of the system is shown in \cref{fig:Architecture}. To be able to adapt quickly to new structural concepts, it leverages text as an additional knowledge source by incorporating CLIPSeg. In the way we use it, with structure-related prompts such as Crane, Truck and Wheel, it yields a likelihood that it shows one of these terms for each image pixel. Then, we construct an undirected image-spanning graph representing image locations (SuperPoint points) with associated features that combine CLIPSeg likelihoods and SuperPoint features. This graph represents textually expressible concepts combined with visual features (in its nodes) and geometric adjacencies in terms of the graph's edges instead of the exact image locations. The graph's nodes are then classified into the part labels by the only trained network of this system, which is a node-classifying GNN.
SAM cannot segment machine parts solely from image data, it requires guidance through bounding boxes or keypoints. Hence, our novel automatic prompt engineering procedure combines the image locations of nodes that share the same part label to query SAM for these machine parts. Finally, SAM produces precise image segmentations for each part.
The rationale behind this graph-based machine learning approach is that it is permutation-invariant but adjacency-preserving.
This means that even if parts of the crane move for similar viewpoints, their graph representations will stay almost the same.
We, therefore, exploit the system-inherent symmetries, which reduce data requirements and training time \cite{bronstein2021geometric}. With our system, a novel few-shot segmentation job can be trained in $<3$ minutes on consumer hardware.

The main contributions of this work are\footnote{Code and sample data are available for academic use on \href{https://github.com/AIT-Assistive-Autonomous-Systems/Hopomop}{GitHub}.}: 
\textbf{(a)} A novel flexible few-shot learning architecture combining CLIPSeg, SuperPoint, and SAM with a custom few-shot trainable GNN for structure-informed part segmentation. \textbf{(b)} A novel SAM prompt engineering process converting a classified image-spanning graph into SAM prompts to robustly generate image part segmentations. \textbf{(c)} A synthetic training data generation pipeline for a truck-mounted loading crane. \textbf{(d)} A readily trained network for crane and truck part detection.

\cref{sec:rel_work} gives a detailed overview of the related work, \cref{sec:data} describes the synthetic data generation process, \cref{sec:methodology} describes all previously mentioned functional blocks and their combination. In \cref{sec:experiments} the system is evaluated on synthetic and real-world data of a truck-mounted loading crane and the DAVIS 2017 dataset.

\section{Related work}
\label{sec:rel_work}
Semantic image segmentation, the task of assigning a class label to each pixel in an image, is a fundamental problem in computer vision. Despite its importance, many applications lack access to adequate datasets with pixel-level annotations, which are needed to train state-of-the-art models. These models include encoder-decoder architectures like U-Net\cite{ronneberger2015u} and SegNet\cite{7803544} or transformer-based ViT\cite{dosovitskiy2020image} and Swin\cite{liu2021swin} architectures. 
One way to alleviate this data problem is Transfer Learning, where a model is first trained on a large dataset and then fine-tuned episodically on a smaller dataset. However, the performance of Transfer Learning highly depends on the pre-trained dataset and the similarity of the tasks\cite{poudel2023exploring, catalano2023few}. The task of learning with an even smaller amount of samples is known as few-shot learning and mostly incorporates meta-learning\cite{vettoruzzo2024advances, SUN202183, nichol2018first}, which is mainly referred to as learning to learn. However, these methods often struggle with high computational costs and the need for numerous training tasks, limiting their practicality for few-shot segmentation. To address these challenges, prototypical learning\cite{catalano2023few} has emerged as a promising alternative, offering a more efficient approach by leveraging prototype representations of classes. However, to extract semantic segmentation masks, the prototype network has to be extended to a pixel-wise level. Superpixels (see \cite{achanta2012slic} for an overview) are a method to partition an image by grouping pixels.
The approach laid out in this work takes a different angle (inspired by \cite{liu2020part}), we first use the lightweight interest points SuperPoint\cite{detone2018superpoint} that are mostly independent of lighting and perspective, and then, we build an image-spanning graph from it. And we do not cluster the image but the graph. 
The image-spanning graph representation enables deep spatial understanding and the usage of efficient graph neural networks\cite{aflalo2023deepcut}. Graph neural networks\cite{Khemani2024} operate on graph-structured data, such as social networks or molecular structures\cite{9046288}. Leaving the usual grid-based image structure by representing the image as a graph and classifying its nodes instead of dense pixel-wise classification is a promising approach for few-shot semantic segmentation\cite{liu2020part}.
The rapid development of large-language model-based approaches such as CLIPSeg\cite{luddecke2022image}, GroundingDINO\cite{liu2023grounding} and OWLV2\cite{minderer2024scaling} that incorporate text prompts for segmentation tasks has led to new opportunities in few- or even zero-shot segmentation. GroundedSAM\cite{ren2024grounded} builds upon GroundingDINO for generating pixel-wise segmentation with SAM\cite{kirillov2023segment} from text prompts. Nevertheless, text-to-segmentation methods still struggle with identifying industrial and robotic parts due to their complex structure and technical naming. Using SAM as the final pixel-wise segmentation step, as SAM-PT\cite{rajivc2023segment} does, allows for high-quality segmentation from sparse point-wise information. By combining different ideas and methodologies, such as interest point detection, graph neural networks, and pre-trained foundation models like CLIPSeg and SAM, we propose a novel approach for few-shot semantic segmentation.

\section{Synthetic data generation}
\label{sec:data}
\begin{figure}[t]
  \centering
   \includegraphics[width=1\linewidth]{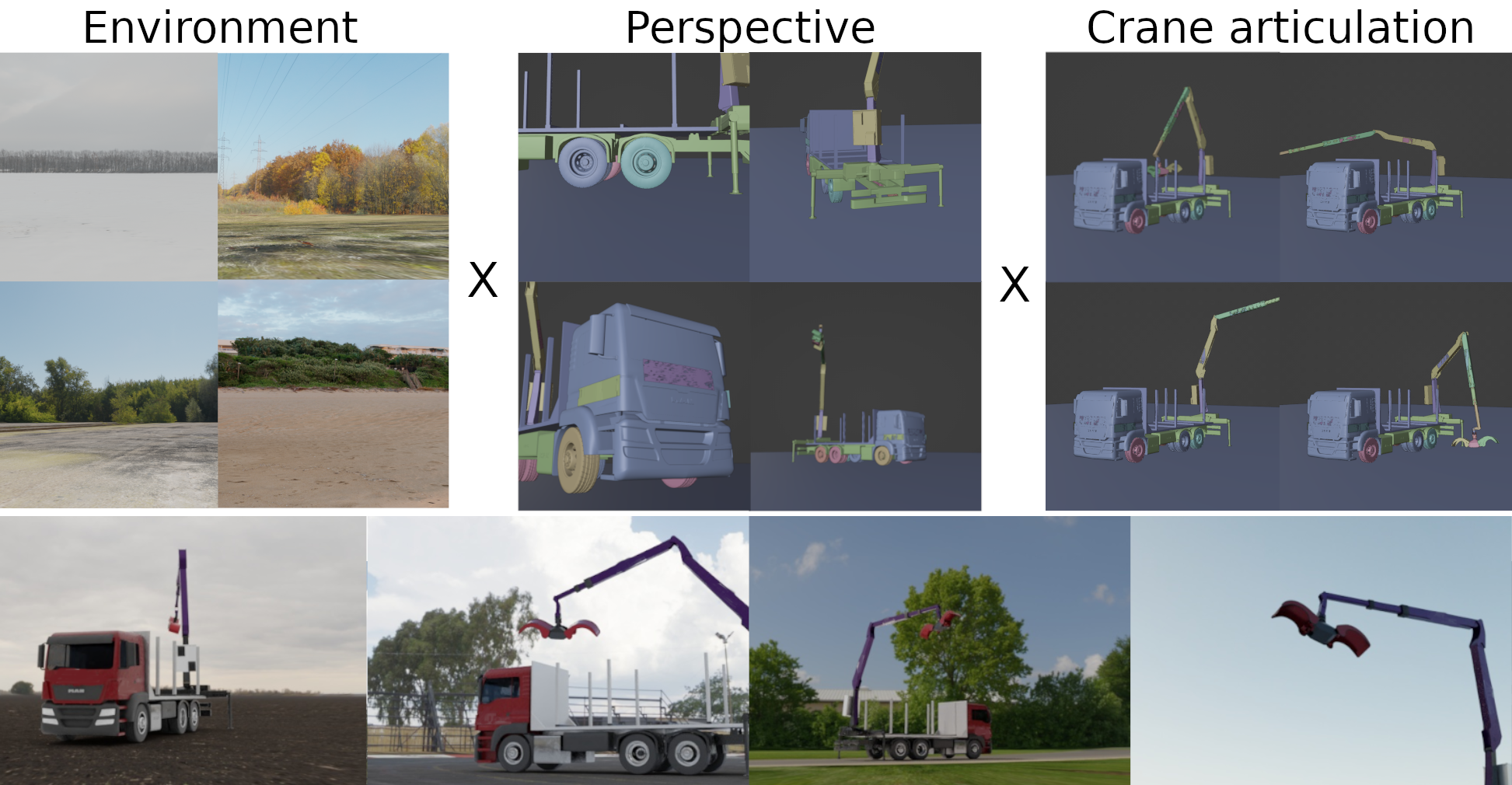}

   \caption{Top: The three axis of domain randomization: environment HDRi maps, changing camera perspective, and different crane arm articulations. Bottom: Samples of the dataset.}
   \label{fig:domain_randomzation}
\end{figure}

\begin{figure}[h]
  \centering
   \includegraphics[width=1\linewidth]{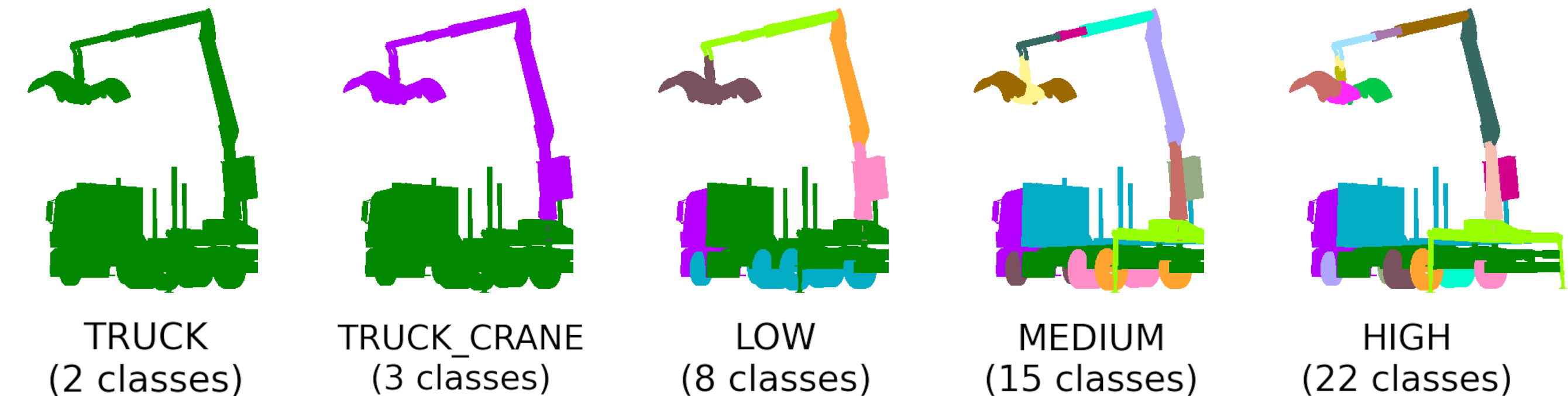}

   \caption{Five different annotation granularity levels, ranging from background and truck differentiation up to 22 individual parts.}
   \label{fig:granularities}
\end{figure}

To provide insights into the segmentation performance of the proposed method, a fully synthetic dataset of a truck-mounted loading crane was created. We used Blender to generate the data, which allows broad domain randomization by creating varying backgrounds, lighting conditions, camera perspectives, and crane arm poses. The main advantage of synthetic data generation is the ability to generate large datasets with labeled data at higher accuracy and a fraction of the cost of manual labeling. Since the model is trained exclusively in a few-shot manner, we cannot rely on standard test-train splits. The training set consists of 100 synthetic renderings, including annotation. During training, we utilize a maximum of 25 random samples, which we consider enough to show the functionality and performance of our approach. The test set consists of 250 samples, introducing five additional lighting environments during generation to increase the diversity of the test data compared to the training data. More details on the training and testing setup are provided in \cref{sec:experiments}. \cref{fig:domain_randomzation} illustrates how lighting environments, camera perspective, and crane arm articulation are combined to create authentic renderings with high diversity. The four rendered samples (bottom) give some intuition about the variety in the dataset. For the performance evaluation of the model, we export five labeled segmentation masks, containing varying levels of granularity ranging from the entire truck and crane to twenty-two individual parts of the crane. \cref{fig:granularities} lists all granularity levels that can be extracted for each sample, beginning with the \textit{Truck} and \textit{Truck Crane} granularity, which distinguishes between background, truck, and crane arm, and progressing to \textit{High} granularity, which includes 22 segmentation classes that segment almost every part of the truck-mounted loading crane. The \textit{Low} granularity is of special interest because it consists of eight challenging, semantically distinct classes, such as wheels, cabin, or loading platform.

\section{Methodology}
\label{sec:methodology}

An overall description of models architecture accompanied by \cref{fig:Architecture} can be found in \cref{sec:intro}. In the following, we describe the used modules in more detail. 
\subsection{Image to graph conversion}
To process images through our pipeline, the first step is to transform each image into an undirected graph. The main driver behind using a graph representation is the reduction of the features to a manageable size and the resulting reduction in computational as well as memory complexity. Additionally, due to their permutation invariance\cite{bronstein2021geometric}, graph-based algorithms help us efficiently learn the structure of moving object components. Lastly, using graphs to describe intricate scenes with hierarchical objects or parts that have spatial relationships seems more intuitive and allows for the removal of a great deal of irrelevant information. Important steps are visualized intuitively in \cref{fig:PipelineSteps}.\\

\textbf{Interest point detection}
For every image, we create a set of interest points $I$ using SuperPoint\cite{detone2018superpoint}. Each interest point $I_i = (x_i, y_i, d_i)$ for all $i \in I$ is characterized by its 2D image coordinates $(x_i, y_i) \in X {\times} Y$, with $X$ and $Y$ being the x- and y-coordinates of the image respectively, and a VGG-based descriptor vector $ d_i \in \mathbb{R}^{256}$. The desired number of interest points $|I|$ and image quality can be taken into consideration when adjusting parameters like the non-maximum suppression radius and keypoint confidence threshold.\\
\textbf{Interest point feature enhancement}
CLIPSeg\cite{luddecke2022image} is a foundation model that takes a text prompt and an image as input and outputs a logit map $L_{xy} \in \mathbb{R}$ with the likelihood of the described object being present at each pixel $(x, y) \in X{\times}Y$. Lower CLIPSeg values for background points help the model focus on the segmentation targets during inference and support convergence during training. We normalize $L_{xy}$ with the sigmoid function to obtain $L_{xy}^s \in [0, 1]$. Then, for all interest points $i\in I$, we use $(x_i, y_i)$ to create an enhanced feature vector $d^e_i = [d_i, L^s_{x_i y_i}]$, where $L^s_{x_i y_i}$ is appended to $d_i$. Finally, we can create an enhanced version of $I$ named $I^e$, where $I^e_i = (x_i, y_i, d^e_i)$ for all $i \in I$.\\
\textbf{Graph construction}
Given the enhanced interest points $I^e$, we construct a graph $G=(I^e, E)$, where the set of nodes is $I^e$ and $E$ represents the set of edges. Each node $v \in I^e$ is defined by $I^e_v = (x_v, y_v, d^e_v)$, where $(x_v, y_v)$ are again the 2D image coordinates and $d^e_v \in \mathbb{R}^{257}$ is the enhanced feature vector of the node $v$. The edges are established based on the Euclidean distance between interest points, connecting each node $v \in I^e$ to its $k$ nearest neighbors using its coordinates $(x_v, y_v)$. During training, however, we select $k$ random vertices from the $k+10$ nearest neighbors for each vertex to create $E$. This contributes to the model's robustness and generalization. For all edges $(v, u) \in E$, the edge weight $w_{vu}$ is defined as the Euclidean distance between the descriptor vectors $d^e_v$ and $d^e_u$. The image is now represented as a graph, where edges represent the spatial relationships between the nodes, and nodes store the feature vectors of the interest points.
\subsection{Graph node classification}

\begin{figure}[h]
  \centering
   \includegraphics[width=1\linewidth]{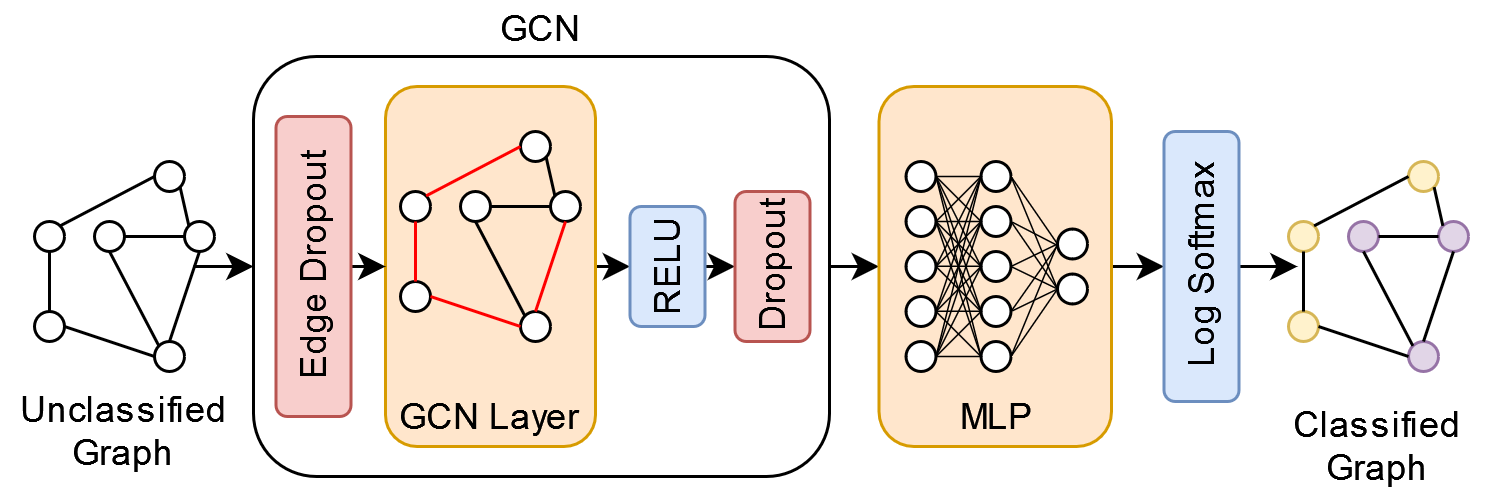}

    \caption{Network architecture of the trainable GCN-based graph classifier.}
    \label{fig:classificationArchiteture}
\end{figure}

\noindent\textbf{Classifier architecture}
As seen in \cref{fig:classificationArchiteture}, the classifier consists of three GCN blocks, followed by a 3-layer perceptron with ReLU activation function for the hidden layer. Since there are numerous GCN layers, we decided to compare GCNConv\cite{kipf2016semi}, GATConv\cite{velivckovic2017graph}, and SAGEConv\cite{hamilton2017inductive} during hyperparameter tuning. Our GCN block consists of an edge dropout layer, the selected GCN layer, a ReLU activation function and feature-level dropout. The three-layer perceptron receives the output from the last GCN block and returns the probability distribution across the classes for each node.\\
\textbf{Training}
For the classifier's training, we obtain the class label for each node $v \in I^e$ from the ground truth segmentation mask via its image coordinate $(x_v,y_v)$. Before the graph creation, the image undergoes spatial augmentation, such as mirroring and random cropping. Furthermore, the graph representation is also augmented by randomly removing edges and nodes and adding noise on node attributes and 2D locations. We train our model using Adam\cite{kingma2014adam} with learning rate scheduling (\textit{StepLR}) and early stopping with the number of train epochs adapted to the class count. Since the background class often has a greater sample size than the other classes, a weighted non-negative log-likelihood loss is applied to account for class imbalance. We calculate the class weights as the inverse of the class frequency. Since we only utilize one or a few samples, we modify the classifier's weights after each image. Moreover, several hyperparameters related to the GCN architecture are defined. These are the \textit{number of graph node neighbors} ($k$), \textit{minimum points} ($|I|$), \textit{non-maximum suppression} (NMS), \textit{SuperPoint threshold} (SPT), and \textit{model type} (MT). The \textit{model type} indicates which GCN layer is used in the classifier (GCN, GAT, or SAGE).
\subsection{Segmentation}
\noindent\textbf{SAM prompt engineering}
Given an input graph, the trained classifier outputs a probability distribution $p_v = (p_{v1}, p_{v2},..., p_{v|C|})$ for each node $v \in I^e$, where $C$ represents the classes and $p_{vc}$ is the probability of the node $v$ belonging to class $c \in C$. The highest probability corresponds to the predicted class of the object represented by the node and its feature vector. SAM\cite{kirillov2023segment} supports two types of input prompts: a set of two-dimensional points accompanied by a class label and a bounding box. Since, dependent on the image, a graph representation may include hundreds or thousands of nodes, we apply the following steps to generate SAM prompts from the classifier's output:
\begin{enumerate}
\item First, the graph nodes are grouped by class: $V_c = \{v \mid \forall v \in I^e, \underset{c \in C}{\arg\max}(p_{vc}) = c\}$.
\item To remove outliers and misclassifications, an isolation forest\cite{4781136} is applied, which results in a reduced set $V^{\prime}_c$ for each class $c \in C$.
\item The isolation forest handles obvious outliers, but a second filter is needed for closer outliers near object boundaries. For each subset $V^{\prime}_c$, we compute the Mahalanobis distance to measures the distance between each point and the entire distribution. This is especially useful for elongated objects, where the distance to the center of the distribution does not accurately represent the distance to the distribution itself. All points above a certain threshold on the normalized Mahalanobis distance are discarded, which leaves us with $V^{\prime\prime}_c$.
\item The remaining nodes $V^{\prime\prime}_c$ are used to construct a bounding box using the minimal and maximal x- and y-coordinates of the associated points for each class $c \in C$.  Similarly, a subset of 2D coordinates are selected from the nodes in $V^{\prime\prime}_c$ for the 2D prompts. To reinforce evenly distributed points, we use farthest point sampling to generate this subset.
\end{enumerate}

\noindent\textbf{SAM segmentation}
The created prompt points have a size of $[B, C, P, 2].$ where $B$ denotes the batch size, $C$ the number of classes, $P$ the number of points per class, and $2$ the 2D coordinates. The bounding box prompts have size $[B,C,4].$ where $4$ denotes the bounding box coordinates. SAM uses the image and the defined prompts as input, and for each class, outputs a triplet of binary masks and a score value ranking the quality of each mask. Eventually, we select the masks with the highest score value as the final segmentation mask for each class. The segmentations are then integrated into a single mask by sorting the binary masks by area and stacking them, beginning with the largest.

\subsection{Implementation}
The proposed method is implemented in PyTorch\cite{paszke2019pytorch} using PyTorch Geometric\cite{fey2019fast} libraries. The transformers\cite{wolf2019huggingface} library provided CLIPSeg ("CIDAS/clipseg-rd64-refined") and SAM ("Facebook/sam-vit-huge"). We obtained the SuperPoint model from the SuperGlue repository\cite{sarlin20superglue}. Every foundation model was used as provided, with no adjustments made.

\subsection{Hyperparameter tuning}
For both the graph node classification and the SAM segmentation part, a random hyperparameter search was performed 20 times on 10 random train and 100 test samples. The most influential hyperparameters for each granularity are shown in \cref{tab:HyperValuesClass} for graph node classification and \cref{tab:HyperValuesSeg} for the final segmentation. All hyperparameters and its definitions can be found in the supplementary material. In the graph node classification stage, we use the F1 metric to assess the influence of the \textit{non-maximum suppression radius} (NR), \textit{maximum points} ($I$), \textit{graph neighbors} ($k$), and \textit{model type} (MT). The top-performing graph convolutional architecture used \textit{SAGEConv} layers and performed well across all levels of granularity. The $k$ value decreased with increasing granularity because finer details rely more on local information, whereas coarser granularities rely on global information. 
 \begin{table}[tb]
  \centering
  {\small{
    \begin{tabular*}{0.9\columnwidth}{l|@{\extracolsep{\stretch{1}}}c@{\extracolsep{\stretch{1}}}c@{\extracolsep{\stretch{1}}}c@{\extracolsep{\stretch{1}}}c|@{\extracolsep{\stretch{1}}}c}
    {Granularity} & {NR} & {$|I|$} & {$k$} & {MT} & {F1}  \\
    \midrule
\textit{Truck} &      4 &      512 &             32 &                 SAGE &      0.91$\pm$0.04  \\
\textit{Truck Crane} &       2 &     1024 &            32 &            SAGE &      0.81$\pm$0.11  \\
\textit{Low}  &     4 &     1024 &             32 &           SAGE &      0.58$\pm$0.16  \\
 \textit{Medium}  &        6 &     1024 &              8 &       SAGE  &  0.39$\pm$0.14 \\
\textit{High} &        4 &      512 &             16 &             SAGE &  0.33$\pm$0.12  \\
    \end{tabular*}
}}
    \caption{Graph node classification part (F1) with NR, $|I|$, $k$, MT hyperparameters for every ganularity.}
    \label{tab:HyperValuesClass}
\end{table}

\cref{tab:HyperValuesSeg} assesses the \textit{SAM prompt type} (SP), \textit{bounding box threshold} (BT), \textit{point threshold} (PT), and \textit{SAM point samples} (SPS) using the dice score. The SP parameter was the most influential hyperparameter, with \textit{Point\&Box} being the leading value, indicating that points and bounding boxes combined achieve the highest performance as SAM input. The SPS parameter follows a similar pattern to $k$, declining with increasing granularity, since smaller regions require fewer points for representation. The bounding box and point threshold settings remain consistent across all granularities. 

\begin{table}[tb]
 \centering
    {\small{  
       \begin{tabular*}{0.9\columnwidth}{l|@{\extracolsep{\stretch{1}}}c@{\extracolsep{\stretch{1}}}c@{\extracolsep{\stretch{1}}}c@{\extracolsep{\stretch{1}}}c|@{\extracolsep{\stretch{1}}}c}
    {Granularity} & {SP} & {BT} & {PT} & {SPS}& {DICE} \\
    \midrule
    \textit{Truck} &             PB &         1.0 &           1.0 &              20 &     0.96$\pm$0.03    \\
    \textit{Truck Crane} &             PB &         1.0 &           0.8 &              15 &     0.85$\pm$0.13  \\
    \textit{Low}  &             PB &         0.8 &           1.0 &              20 &     0.51$\pm$0.14 \\
    \textit{Medium} &             PB &         1.0 &           0.8 &              10 &     0.27$\pm$0.09 \\
    \textit{High} &             PB &         0.8 &           0.8 &              15 &     0.21$\pm$0.07 \\

    \end{tabular*}
    }}
    \caption{Segmentation part (Dice) with SP, BT, PT, SPS hyperparameters for every granularity.}
    \label{tab:HyperValuesSeg}
    \end{table}

\section{Experiments}
\label{sec:experiments}
\subsection{Truck few-shot evaluation}

\begin{figure*}[ht]
  \centering
   \includegraphics[width=0.9\linewidth]{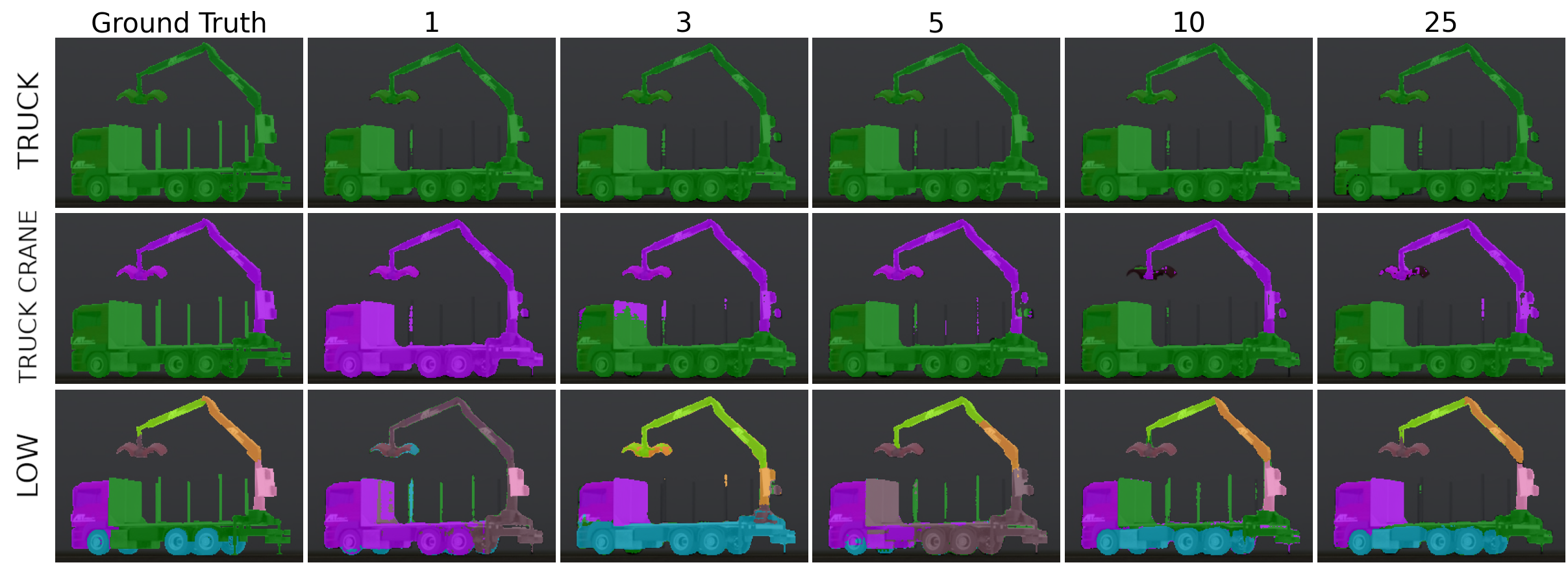}

    \caption{Qualitative segmentation results for different support sample sizes horizontally (1 - 25) vs. structure granularity vertically.}
    \label{fig:sample_comparison}
\end{figure*}

\paragraph{Evaluation}A 5-fold cross-validation is performed on 250 test samples for every granularity and few-shot train sample number ranging from 1 to 25. Hence, we trained a model five times for every granularity on 1, 3, 5, 10, and 25 random samples and evaluated every training run on the same 250 test samples. \cref{tab:truck_results} shows the findings, and the J\&F\cite{pont20172017} metric is employed for assessment of region similarity and contour accuracy. Using just one train sample, lower granularities like \textit{Truck} or \textit{Truck Crane} produces satisfactory results. Ten support samples are required for the \textit{Low} granularity to obtain a J\&F value of 51.1, which is a reasonable result for the eight-class granularity. The J\&F score increases significantly from 1 to 10 sample images, by around 31\% for \textit{Truck Crane} granularity and 63\% for \textit{Low} granularity. For \textit{Medium} and \textit{High} granularities, the results still require improvement even with 25 support samples, which is probably due to the low sample count, high complexity, occlusions, and the interest point detectors' inability to guarantee full point coverage on image regions with small objects or flat surfaces. The segmentation output of our model for the first three granularity levels, trained on 1–25 random support samples from our training set, is displayed in \cref{fig:sample_comparison}. 
To create baselines, we fine-tuned a COCO-pretrained Mask R-CNN \cite{lin2014microsoft, he2017mask} on 15 training and 10 validation samples. The first baseline, \textit{25\textsuperscript{M}} in \cref{tab:truck_results}, directly uses the segmentation masks from Mask R-CNN. In the enhanced baseline (\textit{25\textsuperscript{M+S}}), Mask R-CNN's bounding boxes prompt SAM to generate segmentation masks. Our model outperforms both baselines across all granularity levels. For qualitative comparisons in \cref{tab:truck_results}, see the supplemental material.

\begin{table}[h]
\centering
  {\small{
\begin{tabular}{c|ccccc}

Samples & \textit{Truck} & \textit{Truck Crane} & \textit{Low} & \textit{Medium} & \textit{High} \\
\midrule

1 & 89.2 &  59.4 &  31.4 &  21.1 &  22.0\\
3 & 89.8 &  71.8 &  37.4 &  26.7 &  26.0\\
5 & 89.9 &  75.7 &  41.2 &  30.7 &  28.7\\
10 & 89.9 &  77.8 &  51.1 &  32.8 &  31.2\\
25 & 90.1 &  80.4 &  55.8 &  35.4 &  35.7\\

\midrule

25\textsuperscript{M} & 84.0 & 71.8 & 48.8 & 32.1 & 21.9\\
25\textsuperscript{M+S} & 88.5 & 77.4 & 51.0 & 32.5 & 21.8\\

\end{tabular}
}}
  \caption{Few-shot results of our method with J\&F metric for different granularity levels (columns) and support samples (rows). Baseline evaluation using following alternative methods: Mask R-CNN (25\textsuperscript{M}) and Mask R-CNN with SAM (25\textsuperscript{M+S}). Results are averaged over 5-fold cross-validation on 250 test samples.} 
  \label{tab:truck_results}
\end{table}

\begin{figure}[ht]
  \centering
   \includegraphics[width=0.9\linewidth]{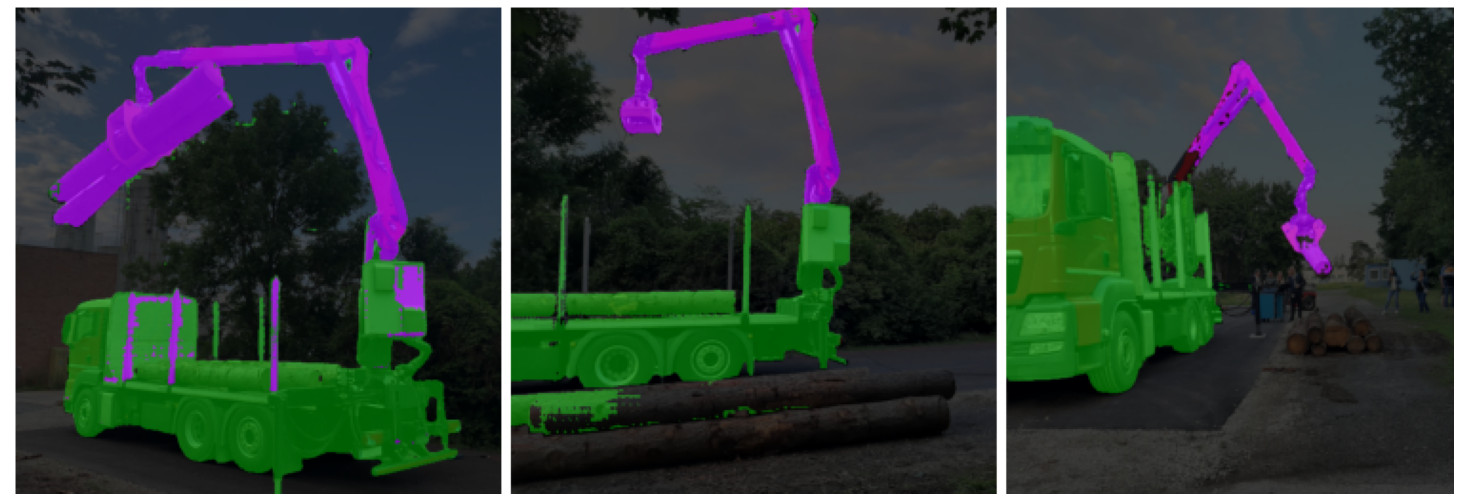}

   \caption{Qualitative results, visualizing the generalization capability of our system trained on synthetic data and applied to real camera images of a truck-mounted loading crane.}
   \label{fig:sim_to_real}
\end{figure}

\paragraph{Synthetic to real}
Since the model was purely trained on a few synthetic samples, we are eager to investigate its generalization capabilities.
Therefore, we curated a small set of real data containing a similar truck-mounted loading crane with manual annotations. 
For this experiment, we trained the model on ten random synthetic samples of our dataset. Even though the real crane arm differs in color and form from its synthetic equivalent from the training data, we can report that our approach can generalize to real data. 
The qualitative results for the \textit{Truck Crane} granularity are shown in \cref{fig:sim_to_real}. The corresponding J\&F scores for the three images from left to right are $65.6$, $65.1$, and $73.7$. For \textit{Truck} granularity, the scores increase to $84.0$, $88.9$, and $92.2$. These generalization results are remarkable, since they allow us to employ this approach in various scenarios where synthetic data is already accessible from 3D models and real data is either unavailable or prohibitively expensive to annotate. 

\subsection{DAVIS 2017 evaluation}
\begin{figure*}
  \centering
   \includegraphics[width=0.9\linewidth]{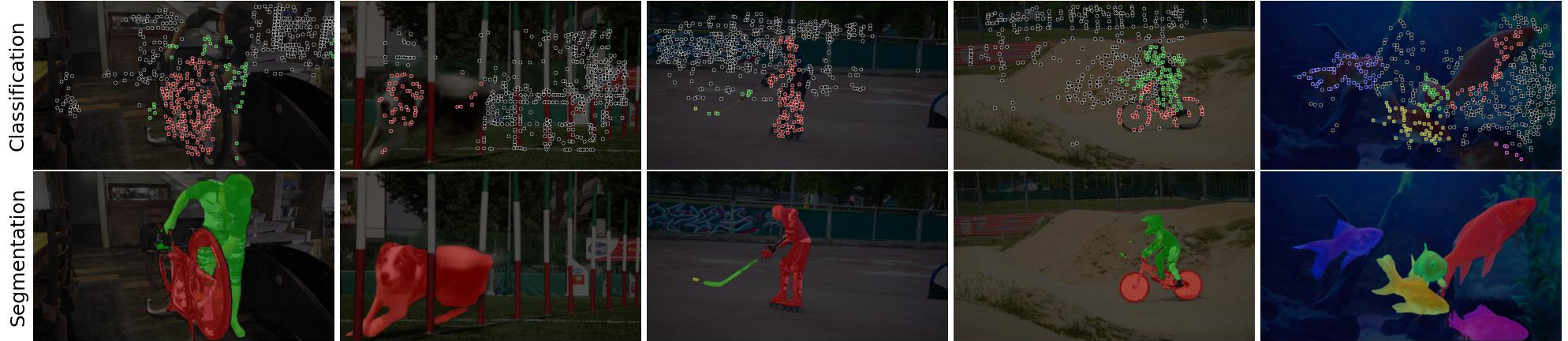}
   \caption{Some qualitative DAVIS 2017 results with our 1-shot model visualizing node classification and segmentation.}
   \label{fig:davis_example}
\end{figure*}

The few-shot segmentation performance of the suggested technique, apart from our particular use case, is benchmarked using the DAVIS 2017\cite{pont20172017} dataset. It is an ideal candidate for evaluation as it contains long, variable sequences with occlusions. First, we train our model on the first frame of the video using the associated ground truth mask. Then, we apply the trained model to the remaining image sequence. The performance was assessed using 30 distinct video sequences from the DAVIS 2017 validation set. We use the $J\&F$ metric with the provided assessment tool set, adhering to the standard evaluation process outlined in \cite{pont20172017}. Our method was designed to be a general-purpose few-shot segmentation method similar to SegGPT\cite{wang2023seggpt} or Painter\cite{wang2023images}, which solely rely on the provided few-shot support samples for training. This is in contrast to video-specific methods like OSMN\cite{yang2018efficient}, OSVOS\cite{caelles2017one}, AGAME\cite{johnander2019generative}, STM\cite{oh2019video}, or XMem\cite{cheng2022xmem}, which make use of temporal information, optical flow, or online learning but operate only on the first (F) frame. 
We used the optimal hyperparameters of the \textit{Low} granularity for training due to the varying class count for different sequences. Then using the first (F), first \& last (FL), and first \& last \& middle (FLM) frames for training. \cref{tab:results_davis} compares the results with several state-of-the-art methodologies. Even though our approach was never intended for such applications, it produces satisfactory results without considering temporal or optical flow information. Furthermore, no fine-tuning is performed before training or during inference, and the foundation models utilized have not been modified at all. During first-frame (F) training, spatial image augmentations are performed to both the first frame and the ground truth mask to improve the model's resilience. Increasing the number of support samples from F to FL or even FLM enhances the results from 54.8 to 65.9 (+20.2\%) and 74.7 (+35.9\%). We consider the findings satisfactory since the main goal was to segment a truck-mounted loading crane and its components using commonly available foundation models without expensive dataset training or curation, and just adapting them to a new assignment. Even with a few support samples, the model can generalize to unknown data. The number of support samples and the granularity of the segmentation challenge both affect the model's performance. \cref{fig:davis_example} displays the qualitative results of our model conditioned on the first images, including a representation of the visualized graph nodes and the final segmentation. Sometimes only one node is needed to define a segmentation, as seen in the \textit{hockey} sample with the yellow class. The \textit{dog} sample shows good occlusion control since the pole is not segmented. 

\begin{table}[th]
\centering
  {\small{
\begin{tabular}{l|ccc}

method & $J\&F$ & $J$ & $F$ \\
\midrule
\multicolumn{4}{c}{General-Purpose Segmentation Methods} \\
Painter \cite{wang2023images}  & 34.6 & 28.5 & 40.8  \\
\textbf{Ours(F)} &54.5 &  50.8 & 58.3 \\
\textbf{Ours(FL)} &65.9 & 62.9 & 68.9 \\
\textbf{Ours(FLM)} &74.5 & 71.5 & 77.6 \\
SegGPT \cite{wang2023seggpt} & 75.6& 72.5 & 78.6 \\
\midrule
\multicolumn{4}{c}{State-of-the-Art Video Segmentation Methods} \\

OSMN \cite{yang2018efficient}& 54.8 & 52.5  & 57.1 \\

OSVOS \cite{caelles2017one} & 60.2 & 56.6 & 63.9 \\
AGAME(+YV) \cite{johnander2019generative} & 69.9 & 67.2 & 72.7 \\
STM \cite{oh2019video}& 81.8 & 79.2 & 84.3 \\
XMem \cite{cheng2022xmem}& 87.7  & 84.0 & 91.4 \\

\end{tabular}
}}
  \caption{Quantitative semi-supervised video segmentation results from the DAVIS 2017 dataset compared to state-of-the-art video segmentation methods.}
  \label{tab:results_davis}
\end{table}

\subsection{Ablation studies}
To better understand the impact of CLIPSeg on segmentation performance, we re-evaluated the DAVIS 2017 dataset, presenting the first frame (F setting) of the sequence with and without feature enhancement. If CLIPSeg is utilized, we use the DAVIS sequence identifier (boat, goldfish, etc.) as the input text prompt. As shown in \cref{tab:abalation_study}, the $J\&F$ scores improve by almost 5 points, indicating that CLIPSeg is incorporating more semantic information into the graph structure, which is good for graph node classification. However, depending on the task, the chosen text prompt can have a substantial impact on segmentation performance. Another significant factor is the \textit{SAM prompt type} (\textit{point}, \textit{box}, \textit{point\&box}). Using a collection of points only to segment objects with holes or occlusions improves the $J\&F$ score by approximately 6 points. In addition to prompt type, the number of points influences segmentation performance as well. Using 25 points instead of 10 results in lower $J\&F$ ratings, although more points provide more descriptive information. This could be due to misclassification in the previous stage. Notably, results can range from 48.7 (worst) to 54.5 (best) based on parameter choice.

\begin{table}[th]
\centering
  {\small{
\begin{tabular}{ccc|c}
\multicolumn{4}{c}{\textbf{DAVIS 2017 - VAL}} \\

  CLIPSeg &SAM Prompt & SAM Points & $J\&F$ \\
\midrule
 \checkmark & points+boxes & 10 &  48.7 \\
 -  & points & 10 & 50.5 \\
 \checkmark & points & 25 & 51.8 \\
 \checkmark & points & 10 & 54.5 \\

\end{tabular}
}}
  \caption{Ablation study about the influence of CLIPSeg feature enhancement, SAM prompt types and SAM point sample count evaluated on the first frame.}
  \label{tab:abalation_study}
\end{table}

\subsection{Performance}
The training and inference time for different class counts is shown in \cref{tab:training_inference_time}. Graph node classification inference differs among granularities due to differences in graph structure complexity, as previously indicated. Segmentation time increases slightly with the class count. A few-shot segmentation job with about three classes can be trained in less than three minutes, and inference takes less than a second, allowing rapid reconditioning. All measurements are computed using an NVIDIA RTX 3090.
\begin{table}[th]
  \centering
  {\small{
  \begin{tabular}{ccc|cccc}
         Classes & Epochs & $k$ & Train & Inf\textsubscript{Class} & Inf\textsubscript{Seg} & Inf\textsubscript{Tot} \\
    \midrule
        2 & 500 & 32  & 108.54 & 0.240 & 0.516 & \textbf{0.756} \\
        3 & 750 & 32  & 167.46 & 0.239& 0.549 & \textbf{0.788} \\
        8 & 1200  & 32  & 434.70 & 0.368 & 0.569 & \textbf{0.937} \\
        16 & 1500  & 8  & 163.44 & 0.180 & 0.554 & \textbf{0.734}  \\
        22 & 2500 & 16  & 346.92 & 0.191 & 0.579 & \textbf{0.770}  \\

  \end{tabular}
  }}
  \caption{Training and inference times in seconds for different class counts and graph neighbors. Epochs are adjusted to class count. Inference time is split into node classification and segmentation.}
  \label{tab:training_inference_time}
\end{table}

\section{Discussion}
\label{sec:discussion}
Few-shot segmentation problems might be addressed in various ways, including meta-learning, prototype-based methods, text-to-segmentation models, and graph-based techniques. Nevertheless, these techniques frequently require a significant investment in time, data curation, and processing power or do not cover specific requirements, such as the segmentation of machinery. This paper proposes a novel approach to few-shot segmentation tasks combining foundation models and graph convolutional networks. We apply cutting-edge foundation models, such as CLIPSeg and SAM, for few-shot segmentation tasks since they are highly generalizable to a broad set of problems. Moreover, semantic segmentation becomes less complex and more approachable when reduced from a pixel-level problem to a point-level one. $J\&F$ scores of 90.1 (\textit{Truck}), 80.4 (\textit{Truck Crane}), and 55.8 (\textit{LOW}) are obtained without the need for costly pre-training by utilizing foundation models straight out of the box. Another important aspect of our approach is synthetic to real data generalization. Our approach achieves a $J\&F$ of 92.2 on real data, where the form and color of the crane arm vary. Getting such scores with training times shorter than five minutes represents a key benefit of our method. Without employing temporal information or pre-training on task-specific data, semi-supervised video segmentation results on the DAVIS 2017 dataset achieve a $J\&F$ score of 71.5 using three support samples, showing a promising result, especially since our approach was not designed for this discipline. 
However, the reliance on interest points, which do not ensure perfect coverage on flat surfaces or in low-contrast areas, sometimes puts our approach at a disadvantage. Furthermore, we still rely on non-deterministic point sampling methods for the SAM input selection. A possible enhancement could be a learning-based approach for that purpose.

\section{Conclusion}
\label{sec:conclusion}
This paper introduced a novel combination of the well-known foundation models CLIPSeg, SAM, and the interest point detector SuperPoint, interfaced with a graph convolutional network for few-shot segmentation tasks, which was tested on images of a truck-mounted loading crane. Our technique can segment unseen data at various degrees of granularity using a few annotated samples, producing remarkable results even with eight different segmentation classes and ten support samples. Lower degrees of detail already generate quality results with only one to three support samples. Aside from that, our model can generalize effectively to real data after being only trained on a synthetic dataset. Another distinguishing characteristic of our approach is that it can be trained on any segmentation task in minutes using only a few annotated examples. The ever-changing nature of machinery and infrastructure necessitates a flexible and quick response to new tasks. Our technique can do so without requiring costly pre-training on large datasets. Using the sheer knowledge provided by specialist foundation models and orchestrating them using a graph representation, an intuitive way to represent machinery or any scene in general, we achieve performant adaptation on unseen data, distinguishing our approach from the others compared.

{\small
\bibliographystyle{ieee_fullname}
\bibliography{egbib}
}

\setcounter{section}{0} 

\onecolumn

\begin{center}
    \Large
    \textbf{Few-shot Structure-Informed Machinery Part Segmentation with Foundation Models and Graph Neural Networks} \\
    \vspace{0.7em}
    \textit{Supplementary Material} \\
    \vspace{3em}
\end{center}

\appendix
\setcounter{page}{1}

\section{Hyperparameters} \label{apx:hyperparamter}

\begin{table*}[!htb]
    \centering
  {\small{
\begin{tabular}{l|c|p{0.7\linewidth}}

\textbf{Hyperparameter} & \textbf{Short} & \textbf{Description}   \\
\hline
NMS Radius & NR & SuperPoint non-maximum-supresion box radius for merging close points. Range: [2 - 6]\\\hline
Min Points & $I$ & Minumum amount of generated interest points for an image. Range: [512 - 2048]\\\hline
Graph Neighbors & $k$ & Amount of neighbor connections for each node. Range: [8 - 32] \\\hline
Hidden Layers & HD & Depth of Hidden Layers in the Graph Classifier. Range: [256 - 1024] \\\hline
Integration Layers & ID & Depth of Integration Layers in the Graph Classifier. Range: [128 - 512]\\\hline
SP Threshold & SPT & SuperPoint threshold value for each interest point quality score. Range: [1.0e\textsuperscript{-04} - 5.0e\textsuperscript{-04}]\\\hline
Model Type & MT & Graph convolution layer type. Values: [GAT, GAN, SAGE] \\\hline
Dropout & DR & Dropout value for the Graph Classifier. Range: [0.1 - 0.3] \\\hline
Dropout Edge & DRE & Edge Dropout value for the Graph Classifier. Range: 0.3 - 0.8]\\\hline
SAM Prompt & SP & SAM prompt type for the Segmentation part. Values: [Point(P), Box(B), Point Box(PB)] \\\hline
Point Threshold & PT & Threshold values for the point's Mahalanobis distance. Range: [0.6 - 1.0] \\\hline
Box Threshold & BT & Threshold values for the bounding boxes' Mahalanobis distance. Range: [0.6 - 1.0] \\\hline
SAM point samples & SPS & SAM input point amount for each class: Range [5 - 20] \\

\end{tabular}
}}
\caption{List of tunable hyperparameters, including description and ranges.}
    \label{tab:HyperparametersList}
\end{table*}

\begin{figure*}[!htb]
    \centering
    \includegraphics[width=0.62\linewidth]{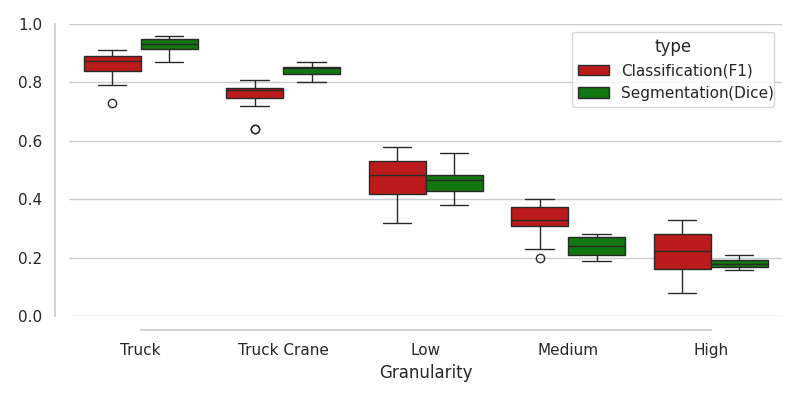}
    \caption{Boxplot visualization of all hyperparameter tuning runs, demonstrating the slight but noticeable differences between various parameter selections at each granularity level. The graph node classification F1 performance is represented in red, and the segmentation Dice score is displayed in green.}
    \label{fig:enter-label}
\end{figure*}

\clearpage
\subsection{Hyperparameter Tuning Classification Part}
    \renewcommand{\arraystretch}{0.6}
    \newcolumntype{?}{!{\vrule width 2pt}}

    \begin{table}[!htb]
    \centering
    \begin{adjustbox}{height=0.25\linewidth}
    \begin{tabular}{|c|c|c|c|c|c|c|c|c|c?c|}
    \toprule

    \multicolumn{10}{|c?}{\textbf{Hyperparameters}} & \multicolumn{1}{c|}{\textbf{Metrics}} \\

    \midrule

    \textbf{RUN} & \textbf{NR} & \textbf{\textit{I}} & \textbf{\textit{k}} & \textbf{HD} & \textbf{ID} & \textbf{SPT} & \textbf{MT} &  \textbf{DR} & \textbf{DRE} & \textbf{F1\textsubscript{M}} \\
    \midrule
    0  &        $2$ &      $512$ &              $8$ &     $1024$ &           $128$ &         $1.0e^{-04}$ &        GNN &       $0.2$ &            $0.5$ &  $0.85\pm0.05$ \\
    1  &        $4$ &     $2048$ &              $8$ &      $512$ &           $256$ &         $2.0e^{-04}$ &       SAGE &       $0.2$ &            $0.3$ &  $0.91\pm0.04$ \\
    2  &        $2$ &     $1024$ &             $32$ &      $256$ &           $512$ &         $2.0e^{-04}$ &        GAT &       $0.1$ &            $0.8$ &  $0.84\pm0.06$ \\\rowcolor{green!20}
    3  &        $4$ &      $512$ &             $32$ &      $512$ &           $128$ &         $1.0e^{-04}$ &       SAGE &       $0.2$ &            $0.8$ &  $0.91\pm0.04$ \\
    4  &        $2$ &      $512$ &              $8$ &     $1024$ &           $128$ &         $1.0e^{-04}$ &       SAGE &       $0.1$ &            $0.8$ &  $0.89\pm0.04$ \\
    5  &        $6$ &      $512$ &             $16$ &      $512$ &           $128$ &         $1.0e^{-04}$ &        GAT &       $0.1$ &            $0.5$ &  $0.84\pm0.06$ \\
    6  &        $4$ &     $1024$ &             $32$ &     $1024$ &           $128$ &         $1.0e^{-04}$ &        GNN &       $0.2$ &            $0.8$ &  $0.81\pm0.04$ \\
    7  &        $2$ &     $1024$ &             $16$ &     $1024$ &           $128$ &         $2.0e^{-04}$ &       SAGE &       $0.3$ &            $0.5$ &  $0.89\pm0.04$ \\
    8  &        $2$ &     $1024$ &              $8$ &      $512$ &           $128$ &         $5.0e^{-04}$ &        GNN &       $0.2$ &            $0.8$ &  $0.87\pm0.04$ \\
    9  &        $4$ &     $1024$ &             $16$ &     $1024$ &           $512$ &         $5.0e^{-04}$ &        GAT &       $0.3$ &            $0.3$ &  $0.83\pm0.07$ \\
    10 &        $2$ &     $2048$ &             $16$ &     $1024$ &           $128$ &         $1.0e^{-04}$ &       SAGE &       $0.3$ &            $0.3$ &   $0.9\pm0.04$ \\
    11 &        $6$ &     $2048$ &             $32$ &      $512$ &           $512$ &         $1.0e^{-04}$ &        GNN &       $0.2$ &            $0.5$ &  $0.87\pm0.05$ \\
    12 &        $2$ &     $1024$ &              $8$ &      $512$ &           $512$ &         $2.0e^{-04}$ &        GAT &       $0.1$ &            $0.5$ &  $0.87\pm0.05$ \\
    13 &        $4$ &     $1024$ &              $8$ &     $1024$ &           $512$ &         $1.0e^{-04}$ &        GNN &       $0.3$ &            $0.5$ &  $0.88\pm0.04$ \\\rowcolor{red!20}
    14 &        $6$ &     $1024$ &             $32$ &      $256$ &           $128$ &         $1.0e^{-04}$ &        GAT &       $0.2$ &            $0.8$ &  $0.73\pm0.07$ \\
    15 &        $4$ &      $512$ &              $8$ &     $1024$ &           $256$ &         $1.0e^{-04}$ &       SAGE &       $0.1$ &            $0.8$ &  $0.89\pm0.05$ \\
    16 &        $6$ &     $2048$ &              $8$ &      $512$ &           $128$ &         $5.0e^{-04}$ &        GAT &       $0.3$ &            $0.3$ &  $0.88\pm0.05$ \\
    17 &        $6$ &     $2048$ &             $16$ &     $1024$ &           $256$ &         $2.0e^{-04}$ &        GNN &       $0.3$ &            $0.3$ &  $0.88\pm0.04$ \\
    18 &        $2$ &      $512$ &             $32$ &      $512$ &           $512$ &         $5.0e^{-04}$ &        GAT &       $0.3$ &            $0.3$ &  $0.79\pm0.06$ \\
    19 &        $2$ &     $1024$ &             $32$ &      $256$ &           $256$ &         $1.0e^{-04}$ &       SAGE &       $0.1$ &            $0.8$ &   $0.9\pm0.04$ \\
    \bottomrule
    \end{tabular}

    \end{adjustbox}
    \caption{20 runs of graph classification hyperparameter tuning with \textbf{\textit{TRUCK}} granularity. Best and worst runs are highlighted in green and red. Metric evaluated on 250 test samples(mean/std).}
    \label{tab:hyp_truck}
    \end{table}

    \begin{table}[!htb]
    \centering
    \begin{adjustbox}{height=0.25\linewidth}
    \begin{tabular}{|c|c|c|c|c|c|c|c|c|c?c|}
    \toprule
    \multicolumn{10}{|c?}{\textbf{Hyperparameters}} & \multicolumn{1}{c|}{\textbf{Metrics}} \\

    \midrule

    \textbf{RUN} & \textbf{NR} & \textbf{\textit{I}} & \textbf{\textit{k}} & \textbf{HD} & \textbf{ID} & \textbf{SPT} & \textbf{MT} &  \textbf{DR} & \textbf{DRE} &  \textbf{F1\textsubscript{M}} \\
    \midrule
    0  &        $6$ &      $512$ &             $16$ &      $512$ &           $256$ &         $5.0e^{-04}$ &        GNN &       $0.2$ &            $0.8$ &  $0.74\pm0.12$ \\
    1  &        $6$ &     $1024$ &             $16$ &      $512$ &           $512$ &         $2.0e^{-04}$ &        GNN &       $0.2$ &            $0.3$ &  $0.77\pm0.12$ \\
    2  &        $6$ &      $512$ &             $32$ &      $512$ &           $128$ &         $5.0e^{-04}$ &       SAGE &       $0.1$ &            $0.3$ &  $0.81\pm0.11$ \\
    3  &        $2$ &      $512$ &             $32$ &      $512$ &           $128$ &         $1.0e^{-04}$ &       SAGE &       $0.1$ &            $0.8$ &   $0.8\pm0.11$ \\
    4  &        $4$ &     $2048$ &             $32$ &      $256$ &           $128$ &         $2.0e^{-04}$ &        GNN &       $0.2$ &            $0.5$ &  $0.76\pm0.12$ \\
    5  &        $2$ &     $1024$ &              $8$ &      $256$ &           $256$ &         $1.0e^{-04}$ &        GAT &       $0.1$ &            $0.5$ &  $0.78\pm0.09$ \\
    6  &        $6$ &     $2048$ &              $8$ &      $512$ &           $512$ &         $5.0e^{-04}$ &        GAT &       $0.3$ &            $0.8$ &   $0.78\pm0.1$ \\
    7  &        $2$ &     $2048$ &             $32$ &      $256$ &           $512$ &         $5.0e^{-04}$ &        GNN &       $0.2$ &            $0.5$ &  $0.76\pm0.11$ \\
    8  &        $2$ &      $512$ &             $32$ &     $1024$ &           $128$ &         $2.0e^{-04}$ &        GAT &       $0.3$ &            $0.8$ &   $0.64\pm0.1$ \\\rowcolor{green!20}
    9  &        $2$ &     $1024$ &             $32$ &     $1024$ &           $512$ &         $2.0e^{-04}$ &       SAGE &       $0.2$ &            $0.3$ &  $0.81\pm0.11$ \\
    10 &        $2$ &     $1024$ &              $8$ &      $256$ &           $512$ &         $2.0e^{-04}$ &       SAGE &       $0.1$ &            $0.8$ &   $0.78\pm0.1$ \\
    11 &        $2$ &     $2048$ &              $8$ &     $1024$ &           $512$ &         $1.0e^{-04}$ &       SAGE &       $0.1$ &            $0.5$ &   $0.79\pm0.1$ \\
    12 &        $4$ &     $1024$ &              $8$ &      $256$ &           $128$ &         $5.0e^{-04}$ &        GNN &       $0.2$ &            $0.3$ &  $0.78\pm0.11$ \\
    13 &        $6$ &      $512$ &             $16$ &     $1024$ &           $256$ &         $1.0e^{-04}$ &        GNN &       $0.1$ &            $0.8$ &  $0.72\pm0.11$ \\
    14 &        $4$ &      $512$ &             $32$ &      $256$ &           $256$ &         $5.0e^{-04}$ &       SAGE &       $0.1$ &            $0.5$ &  $0.73\pm0.12$ \\
    15 &        $4$ &     $1024$ &              $8$ &      $512$ &           $128$ &         $2.0e^{-04}$ &        GNN &       $0.1$ &            $0.5$ &   $0.78\pm0.1$ \\
    16 &        $4$ &     $1024$ &              $8$ &      $512$ &           $512$ &         $5.0e^{-04}$ &        GAT &       $0.1$ &            $0.5$ &   $0.75\pm0.1$ \\\rowcolor{red!20}
    17 &        $6$ &      $512$ &             $16$ &      $512$ &           $256$ &         $2.0e^{-04}$ &        GAT &       $0.3$ &            $0.5$ &  $0.64\pm0.12$ \\
    18 &        $6$ &      $512$ &              $8$ &      $256$ &           $512$ &         $5.0e^{-04}$ &        GNN &       $0.3$ &            $0.5$ &  $0.77\pm0.12$ \\
    19 &        $2$ &     $1024$ &             $16$ &     $1024$ &           $128$ &         $5.0e^{-04}$ &       SAGE &       $0.1$ &            $0.8$ &    $0.8\pm0.1$ \\
    \bottomrule
    \end{tabular}
    \end{adjustbox}
    \caption{20 runs of graph classification hyperparameter tuning with \textbf{\textit{TRUCK CRANE}} granularity. Best and worst runs are highlighted in green and red. Metric evaluated on 250 test samples(mean/std).}
    \label{tab:hyp_truck_crane}
    \end{table}

    \begin{table}[!htbp]
    \centering
    \begin{adjustbox}{height=0.25\linewidth}
    \begin{tabular}{|c|c|c|c|c|c|c|c|c|c?c|}
    \toprule
    \multicolumn{10}{|c?}{\textbf{Hyperparameters}} & \multicolumn{1}{c|}{\textbf{Metrics}} \\

    \midrule

    \textbf{RUN} & \textbf{NR} & \textbf{\textit{I}} & \textbf{\textit{k}} & \textbf{HD} & \textbf{ID} & \textbf{SPT} & \textbf{MT} &  \textbf{DR} & \textbf{DRE} &  \textbf{F1\textsubscript{M}} \\
    \midrule
    0  &        $4$ &     $1024$ &             $32$ &     $1024$ &           $256$ &         $1.0e^{-04}$ &        GAT &       $0.1$ &            $0.3$ &  $0.32\pm0.09$ \\
    1  &        $6$ &     $2048$ &              $8$ &      $256$ &           $128$ &         $5.0e^{-04}$ &        GNN &       $0.2$ &            $0.5$ &  $0.43\pm0.12$ \\
    2  &        $2$ &     $1024$ &              $8$ &     $1024$ &           $256$ &         $5.0e^{-04}$ &        GNN &       $0.3$ &            $0.8$ &  $0.53\pm0.15$ \\\rowcolor{green!20}
    3  &        $4$ &     $1024$ &             $32$ &      $512$ &           $256$ &         $5.0e^{-04}$ &       SAGE &       $0.2$ &            $0.3$ &  $0.58\pm0.16$ \\
    4  &        $2$ &     $2048$ &             $16$ &      $256$ &           $256$ &         $5.0e^{-04}$ &       SAGE &       $0.3$ &            $0.8$ &  $0.53\pm0.16$ \\
    5  &        $6$ &     $1024$ &             $16$ &     $1024$ &           $256$ &         $5.0e^{-04}$ &        GNN &       $0.1$ &            $0.5$ &  $0.42\pm0.12$ \\
    6  &        $2$ &     $2048$ &             $16$ &      $256$ &           $256$ &         $2.0e^{-04}$ &        GAT &       $0.1$ &            $0.8$ &   $0.5\pm0.14$ \\
    7  &        $4$ &      $512$ &             $16$ &      $256$ &           $128$ &         $2.0e^{-04}$ &        GAT &       $0.3$ &            $0.5$ &  $0.42\pm0.12$ \\
    8  &        $6$ &     $1024$ &             $32$ &      $512$ &           $128$ &         $1.0e^{-04}$ &        GNN &       $0.2$ &            $0.3$ &  $0.44\pm0.12$ \\
    9  &        $4$ &     $2048$ &              $8$ &      $512$ &           $256$ &         $2.0e^{-04}$ &       SAGE &       $0.3$ &            $0.3$ &  $0.54\pm0.16$ \\
    10 &        $4$ &      $512$ &             $32$ &      $256$ &           $256$ &         $5.0e^{-04}$ &        GNN &       $0.1$ &            $0.8$ &  $0.37\pm0.11$ \\
    11 &        $4$ &      $512$ &              $8$ &      $256$ &           $512$ &         $1.0e^{-04}$ &        GAT &       $0.2$ &            $0.8$ &  $0.36\pm0.11$ \\
    12 &        $2$ &     $1024$ &             $16$ &      $512$ &           $512$ &         $2.0e^{-04}$ &        GNN &       $0.1$ &            $0.8$ &  $0.53\pm0.15$ \\\rowcolor{red!20}
    13 &        $6$ &      $512$ &              $8$ &     $1024$ &           $512$ &         $5.0e^{-04}$ &        GAT &       $0.2$ &            $0.3$ &  $0.35\pm0.11$ \\
    14 &        $6$ &      $512$ &              $8$ &      $512$ &           $256$ &         $1.0e^{-04}$ &        GNN &       $0.3$ &            $0.8$ &  $0.48\pm0.13$ \\
    15 &        $6$ &     $1024$ &             $16$ &      $512$ &           $256$ &         $1.0e^{-04}$ &        GNN &       $0.3$ &            $0.5$ &  $0.49\pm0.13$ \\
    16 &        $2$ &     $2048$ &              $8$ &      $256$ &           $128$ &         $1.0e^{-04}$ &        GNN &       $0.1$ &            $0.8$ &  $0.53\pm0.15$ \\
    17 &        $2$ &      $512$ &             $32$ &      $256$ &           $128$ &         $2.0e^{-04}$ &        GAT &       $0.2$ &            $0.5$ &  $0.41\pm0.12$ \\
    18 &        $4$ &     $2048$ &              $8$ &     $1024$ &           $256$ &         $2.0e^{-04}$ &       SAGE &       $0.1$ &            $0.5$ &  $0.53\pm0.15$ \\
    19 &        $2$ &      $512$ &              $8$ &      $512$ &           $256$ &         $1.0e^{-04}$ &       SAGE &       $0.3$ &            $0.5$ &  $0.55\pm0.15$ \\
    \bottomrule
    \end{tabular}
    \end{adjustbox}
    \caption{20 runs of graph classification hyperparameter tuning with \textbf{\textit{LOW}} granularity. Best and worst runs are highlighted in green and red. Metric evaluated on 250 test samples(mean/std).}
    \label{tab:hyp_low}
    \end{table}

    \begin{table}[!htbp]
    \centering
    \begin{adjustbox}{height=0.25\linewidth}
    \begin{tabular}{|c|c|c|c|c|c|c|c|c|c?c|}
    \toprule
    \multicolumn{10}{|c?}{\textbf{Hyperparameters}} & \multicolumn{1}{c|}{\textbf{Metrics}} \\

    \midrule

    \textbf{RUN} & \textbf{NR} & \textbf{\textit{I}} & \textbf{\textit{k}} & \textbf{HD} & \textbf{ID} & \textbf{SPT} & \textbf{MT} &  \textbf{DR} & \textbf{DRE} &  \textbf{F1\textsubscript{M}} \\
    \midrule
    0  &        $6$ &     $2048$ &              $8$ &      $512$ &           $512$ &         $1.0e^{-04}$ &        GAT &       $0.1$ &            $0.5$ &   $0.35\pm0.1$ \\
    1  &        $4$ &     $2048$ &             $32$ &     $1024$ &           $512$ &         $1.0e^{-04}$ &        GNN &       $0.3$ &            $0.3$ &  $0.34\pm0.12$ \\
    2  &        $6$ &      $512$ &              $8$ &     $1024$ &           $256$ &         $1.0e^{-04}$ &       SAGE &       $0.3$ &            $0.5$ &   $0.4\pm0.14$ \\
    3  &        $6$ &     $1024$ &             $32$ &      $256$ &           $128$ &         $2.0e^{-04}$ &        GNN &       $0.1$ &            $0.8$ &  $0.23\pm0.08$ \\
    4  &        $2$ &     $1024$ &             $16$ &      $512$ &           $512$ &         $1.0e^{-04}$ &        GAT &       $0.2$ &            $0.3$ &   $0.31\pm0.1$ \\
    5  &        $2$ &      $512$ &             $16$ &      $256$ &           $256$ &         $5.0e^{-04}$ &       SAGE &       $0.2$ &            $0.5$ &  $0.39\pm0.14$ \\
    6  &        $2$ &     $2048$ &              $8$ &      $256$ &           $256$ &         $1.0e^{-04}$ &        GNN &       $0.3$ &            $0.5$ &  $0.37\pm0.14$ \\
    7  &        $4$ &     $1024$ &              $8$ &      $512$ &           $512$ &         $1.0e^{-04}$ &        GNN &       $0.1$ &            $0.8$ &   $0.33\pm0.1$ \\
    8  &        $2$ &     $1024$ &             $32$ &     $1024$ &           $256$ &         $2.0e^{-04}$ &       SAGE &       $0.3$ &            $0.5$ &   $0.4\pm0.15$ \\
    9  &        $2$ &      $512$ &             $16$ &      $512$ &           $512$ &         $1.0e^{-04}$ &       SAGE &       $0.1$ &            $0.8$ &   $0.32\pm0.1$ \\
    10 &        $4$ &     $1024$ &              $8$ &      $256$ &           $512$ &         $1.0e^{-04}$ &       SAGE &       $0.2$ &            $0.3$ &  $0.37\pm0.13$ \\
    11 &        $6$ &      $512$ &             $16$ &     $1024$ &           $128$ &         $2.0e^{-04}$ &       SAGE &       $0.3$ &            $0.3$ &  $0.33\pm0.11$ \\\rowcolor{green!20}
    12 &        $6$ &     $1024$ &              $8$ &      $512$ &           $128$ &         $1.0e^{-04}$ &       SAGE &       $0.3$ &            $0.8$ &  $0.39\pm0.14$ \\
    13 &        $4$ &     $1024$ &             $16$ &     $1024$ &           $128$ &         $2.0e^{-04}$ &        GNN &       $0.1$ &            $0.3$ &   $0.31\pm0.1$ \\
    14 &        $6$ &     $1024$ &             $16$ &      $256$ &           $512$ &         $5.0e^{-04}$ &        GNN &       $0.3$ &            $0.3$ &  $0.31\pm0.11$ \\
    15 &        $2$ &      $512$ &              $8$ &      $512$ &           $512$ &         $5.0e^{-04}$ &        GNN &       $0.1$ &            $0.5$ &   $0.29\pm0.1$ \\
    16 &        $4$ &      $512$ &              $8$ &      $256$ &           $128$ &         $2.0e^{-04}$ &        GNN &       $0.2$ &            $0.5$ &   $0.31\pm0.1$ \\
    17 &        $6$ &     $1024$ &             $16$ &      $256$ &           $512$ &         $5.0e^{-04}$ &       SAGE &       $0.2$ &            $0.8$ &  $0.28\pm0.08$ \\\rowcolor{red!20}
    18 &        $6$ &     $1024$ &             $16$ &     $1024$ &           $256$ &         $5.0e^{-04}$ &        GAT &       $0.2$ &            $0.8$ &   $0.2\pm0.06$ \\
    19 &        $4$ &      $512$ &              $8$ &     $1024$ &           $512$ &         $1.0e^{-04}$ &       SAGE &       $0.3$ &            $0.8$ &  $0.39\pm0.13$ \\
    \bottomrule
    \end{tabular}
    \end{adjustbox}
    \caption{20 runs of graph classification hyperparameter tuning with \textbf{\textit{MEDIUM}} granularity. Best and worst runs are highlighted in green and red. Metric evaluated on 250 test samples(mean/std).}
    \label{tab:hyp_medium}
    \end{table}

    \begin{table}[!htbp]
    \centering
    \begin{adjustbox}{height=0.25\linewidth}
    \begin{tabular}{|c|c|c|c|c|c|c|c|c|c?c|}
    \toprule
    \multicolumn{10}{|c?}{\textbf{Hyperparameters}} & \multicolumn{1}{c|}{\textbf{Metrics}} \\

    \midrule

    \textbf{RUN} & \textbf{NR} & \textbf{\textit{I}} & \textbf{\textit{k}} & \textbf{HD} & \textbf{ID} & \textbf{SPT} & \textbf{MT} &  \textbf{DR} & \textbf{DRE} &  \textbf{F1\textsubscript{M}} \\
    \midrule
    0  &        $4$ &      $512$ &             $16$ &      $512$ &           $512$ &         $1.0e^{-04}$ &       SAGE &       $0.2$ &            $0.5$ &  $0.32\pm0.13$ \\
    1  &        $2$ &     $2048$ &             $16$ &      $256$ &           $128$ &         $1.0e^{-04}$ &       SAGE &       $0.1$ &            $0.8$ &  $0.29\pm0.11$ \\
    2  &        $6$ &      $512$ &             $32$ &      $512$ &           $256$ &         $1.0e^{-04}$ &        GAT &       $0.3$ &            $0.8$ &   $0.1\pm0.03$ \\
    3  &        $4$ &     $2048$ &              $8$ &      $512$ &           $512$ &         $5.0e^{-04}$ &        GAT &       $0.3$ &            $0.3$ &  $0.27\pm0.09$ \\
    4  &        $6$ &      $512$ &             $16$ &      $256$ &           $512$ &         $2.0e^{-04}$ &        GAT &       $0.1$ &            $0.8$ &  $0.14\pm0.04$ \\
    5  &        $6$ &      $512$ &              $8$ &     $1024$ &           $256$ &         $5.0e^{-04}$ &        GAT &       $0.1$ &            $0.3$ &  $0.13\pm0.04$ \\
    6  &        $2$ &     $2048$ &              $8$ &      $512$ &           $128$ &         $2.0e^{-04}$ &        GNN &       $0.3$ &            $0.5$ &   $0.28\pm0.1$ \\
    7  &        $4$ &      $512$ &              $8$ &      $256$ &           $512$ &         $1.0e^{-04}$ &        GNN &       $0.1$ &            $0.3$ &  $0.23\pm0.07$ \\
    8  &        $4$ &      $512$ &             $32$ &      $256$ &           $128$ &         $2.0e^{-04}$ &        GAT &       $0.2$ &            $0.3$ &  $0.19\pm0.06$ \\
    9  &        $6$ &     $1024$ &             $16$ &      $512$ &           $512$ &         $5.0e^{-04}$ &        GAT &       $0.2$ &            $0.5$ &  $0.22\pm0.07$ \\\rowcolor{red!20}
    10 &        $2$ &      $512$ &             $32$ &     $1024$ &           $512$ &         $1.0e^{-04}$ &        GAT &       $0.1$ &            $0.3$ &  $0.08\pm0.03$ \\
    11 &        $4$ &      $512$ &             $16$ &     $1024$ &           $128$ &         $1.0e^{-04}$ &        GNN &       $0.3$ &            $0.8$ &  $0.17\pm0.06$ \\
    12 &        $2$ &      $512$ &             $16$ &      $256$ &           $256$ &         $5.0e^{-04}$ &        GAT &       $0.3$ &            $0.5$ &  $0.22\pm0.08$ \\
    13 &        $2$ &     $2048$ &              $8$ &     $1024$ &           $512$ &         $2.0e^{-04}$ &        GNN &       $0.2$ &            $0.8$ &  $0.29\pm0.12$ \\\rowcolor{green!20}
    14 &        $4$ &      $512$ &             $16$ &      $256$ &           $256$ &         $1.0e^{-04}$ &       SAGE &       $0.1$ &            $0.5$ &  $0.33\pm0.12$ \\
    15 &        $6$ &      $512$ &             $16$ &     $1024$ &           $512$ &         $5.0e^{-04}$ &       SAGE &       $0.3$ &            $0.5$ &  $0.22\pm0.07$ \\
    16 &        $4$ &     $1024$ &              $8$ &      $256$ &           $128$ &         $2.0e^{-04}$ &        GAT &       $0.3$ &            $0.5$ &  $0.24\pm0.08$ \\
    17 &        $4$ &     $1024$ &              $8$ &     $1024$ &           $256$ &         $1.0e^{-04}$ &       SAGE &       $0.1$ &            $0.8$ &   $0.3\pm0.12$ \\
    18 &        $6$ &     $1024$ &             $32$ &      $512$ &           $128$ &         $1.0e^{-04}$ &        GAT &       $0.1$ &            $0.8$ &  $0.13\pm0.04$ \\
    19 &        $2$ &     $2048$ &              $8$ &      $512$ &           $256$ &         $2.0e^{-04}$ &        GAT &       $0.3$ &            $0.5$ &  $0.27\pm0.09$ \\
    \bottomrule
    \end{tabular}
    \end{adjustbox}
    \caption{20 runs of graph classification hyperparameter tuning with \textbf{\textit{HIGH}} granularity. Best and worst runs are highlighted in green and red. Metric evaluated on 250 test samples(mean/std).}
    \label{tab:hyp_high}
    \end{table}

    \clearpage
\clearpage
\subsection{Hyperparameter Tuning Segmentation Part}
\renewcommand{\arraystretch}{0.6}
\begin{table}[!htb]
\centering
\begin{adjustbox}{height=0.25\linewidth}
\begin{tabular}{|c|c|c|c|c?c|c|}
\toprule
\multicolumn{5}{|c?}{\textbf{Hyperparameters}} & \multicolumn{1}{c|}{\textbf{Metrics}} \\

\midrule

\textbf{RUN} &\textbf{SP} & \textbf{BT} & \textbf{PT} & \textbf{SPS}& \textbf{DICE} \\
\midrule
0  &              P &         $1.0$ &           $0.6$ &              $15$ &     $0.92\pm0.06$  \\
1  &             PB &         $0.8$ &           $0.8$ &              $10$ &     $0.94\pm0.05$  \\
2  &              B &         $0.6$ &           $0.6$ &              $15$ &     $0.87\pm0.09$  \\\rowcolor{green!20}
3  &             PB &         $1.0$ &           $1.0$ &              $20$ &     $0.96\pm0.03$  \\
4  &              P &         $1.0$ &           $0.8$ &              $15$ &     $0.93\pm0.07$  \\
5  &             PB &         $0.6$ &           $0.6$ &              $10$ &      $0.9\pm0.07$  \\
6  &              B &         $1.0$ &           $1.0$ &              $20$ &     $0.96\pm0.03$  \\
7  &              B &         $0.6$ &           $1.0$ &              $20$ &     $0.87\pm0.09$  \\
8  &              B &         $1.0$ &           $1.0$ &              $10$ &     $0.95\pm0.08$  \\
9  &              P &         $1.0$ &           $1.0$ &              $20$ &     $0.95\pm0.03$  \\
10 &             PB &         $1.0$ &           $1.0$ &              $20$ &     $0.96\pm0.03$  \\\rowcolor{red!20}
11 &              B &         $0.6$ &           $1.0$ &              $20$ &      $0.87\pm0.1$  \\
12 &              P &         $1.0$ &           $1.0$ &              $20$ &     $0.94\pm0.06$  \\
13 &             PB &         $0.6$ &           $0.6$ &              $10$ &     $0.89\pm0.07$  \\
14 &              P &         $1.0$ &           $0.6$ &              $15$ &     $0.92\pm0.04$  \\
15 &              P &         $1.0$ &           $0.6$ &              $15$ &     $0.92\pm0.05$  \\
16 &              P &         $0.6$ &           $1.0$ &              $15$ &     $0.95\pm0.03$  \\
17 &             PB &         $0.8$ &           $0.6$ &              $10$ &     $0.94\pm0.04$  \\
18 &              B &         $0.8$ &           $1.0$ &              $20$ &     $0.93\pm0.07$  \\
19 &              P &         $0.8$ &           $0.6$ &              $10$ &     $0.92\pm0.06$  \\
\bottomrule
\end{tabular}
\end{adjustbox}

\caption{20 runs of segmentation hyperparameter tuning with \textbf{\textit{TRUCK}} granularity. Best and worst runs are highlighted in green and red. Metric evaluated on 250 test samples(mean/std).}
\label{tab:hyp_seg_truck}

\end{table}

\begin{table}[!htb]
\centering
\begin{adjustbox}{height=0.25\linewidth}
\begin{tabular}{|c|c|c|c|c?c|c|}
\toprule
\multicolumn{5}{|c?}{\textbf{Hyperparameters}} & \multicolumn{1}{c|}{\textbf{Metrics}} \\

\midrule

\textbf{RUN} & \textbf{SP} & \textbf{BT} & \textbf{PT} & \textbf{SPS}& \textbf{DICE} \\
\midrule
0  &             PB &         $0.8$ &           $0.6$ &              $20$ &     $0.86\pm0.11$  \\
1  &              B &         $0.8$ &           $0.6$ &               $5$ &     $0.85\pm0.13$  \\
2  &             PB &         $1.0$ &           $0.6$ &              $10$ &     $0.86\pm0.11$  \\
3  &             PB &         $1.0$ &           $0.6$ &               $5$ &     $0.86\pm0.12$  \\
4  &             PB &         $1.0$ &           $1.0$ &              $15$ &     $0.85\pm0.14$  \\
5  &              P &         $0.6$ &           $0.8$ &              $20$ &     $0.85\pm0.14$  \\
6  &             PB &         $0.6$ &           $0.6$ &              $15$ &     $0.84\pm0.12$  \\
7  &             PB &         $1.0$ &           $0.8$ &               $5$ &     $0.86\pm0.11$  \\
8  &              P &         $0.8$ &           $0.8$ &               $5$ &      $0.8\pm0.16$  \\
9  &              B &         $0.8$ &           $1.0$ &               $5$ &     $0.83\pm0.14$  \\
10 &              P &         $0.8$ &           $1.0$ &              $15$ &      $0.8\pm0.17$  \\\rowcolor{red!20}
11 &              B &         $0.6$ &           $0.6$ &               $5$ &     $0.81\pm0.14$  \\
12 &             PB &         $0.8$ &           $0.6$ &              $10$ &     $0.87\pm0.11$  \\
13 &              P &         $0.8$ &           $0.8$ &              $20$ &     $0.83\pm0.15$  \\
14 &             PB &         $0.6$ &           $0.8$ &               $5$ &     $0.83\pm0.12$  \\
15 &             PB &         $0.6$ &           $0.8$ &              $15$ &     $0.84\pm0.12$  \\\rowcolor{green!20}
16 &             PB &         $1.0$ &           $0.8$ &              $15$ &     $0.85\pm0.13$  \\
17 &              P &         $1.0$ &           $0.8$ &              $10$ &     $0.85\pm0.14$  \\
18 &             PB &         $1.0$ &           $0.8$ &              $20$ &     $0.85\pm0.12$  \\
19 &             PB &         $1.0$ &           $0.6$ &              $20$ &     $0.85\pm0.13$  \\
\bottomrule
\end{tabular}
\end{adjustbox}

\caption{20 runs of segmentation hyperparameter tuning with \textbf{\textit{TRUCK CRANE}} granularity. Best and worst runs are highlighted in green and red. Metric evaluated on 250 test samples(mean/std).}
\label{tab:hyp_seg_truck_crane}

\end{table}

\begin{table}[!htb]
\centering
\begin{adjustbox}{height=0.25\linewidth}
\begin{tabular}{|c|c|c|c|c?c|c|}
\toprule
\multicolumn{5}{|c?}{\textbf{Hyperparameters}} & \multicolumn{1}{c|}{\textbf{Metrics}} \\

\midrule

\textbf{RUN} & \textbf{SP} & \textbf{BT} & \textbf{PT} & \textbf{SPS}& \textbf{DICE} \\
\midrule
0  &             PB &         $0.8$ &           $1.0$ &              $20$ &     $0.51\pm0.14$ \\\rowcolor{green!20}
1  &             PB &         $0.8$ &           $0.6$ &              $10$ &     $0.52\pm0.14$ \\
2  &              P &         $0.6$ &           $0.6$ &              $15$ &     $0.46\pm0.14$ \\
3  &             PB &         $0.6$ &           $0.6$ &              $10$ &     $0.56\pm0.15$ \\
4  &              P &         $0.6$ &           $0.8$ &              $20$ &     $0.47\pm0.14$ \\
5  &              P &         $1.0$ &           $0.6$ &              $20$ &     $0.49\pm0.14$ \\
6  &              B &         $1.0$ &           $1.0$ &               $5$ &     $0.43\pm0.13$ \\
7  &              P &         $0.8$ &           $0.6$ &               $5$ &      $0.4\pm0.13$ \\
8  &              P &         $0.6$ &           $1.0$ &               $5$ &     $0.38\pm0.12$ \\\rowcolor{red!20}
9  &              P &         $0.8$ &           $0.8$ &              $10$ &     $0.44\pm0.14$ \\
10 &              P &         $0.6$ &           $0.6$ &              $15$ &     $0.47\pm0.14$ \\
11 &              P &         $1.0$ &           $1.0$ &              $10$ &     $0.41\pm0.13$ \\
12 &              P &         $0.8$ &           $0.8$ &              $15$ &     $0.45\pm0.14$ \\
13 &              P &         $1.0$ &           $0.8$ &              $20$ &     $0.47\pm0.15$ \\
14 &             PB &         $1.0$ &           $1.0$ &              $15$ &     $0.48\pm0.13$ \\
15 &              B &         $0.8$ &           $1.0$ &               $5$ &     $0.49\pm0.15$ \\
16 &              P &         $0.8$ &           $0.8$ &              $10$ &     $0.42\pm0.12$ \\
17 &             PB &         $1.0$ &           $1.0$ &              $20$ &     $0.47\pm0.13$ \\
18 &              B &         $1.0$ &           $0.8$ &               $5$ &     $0.43\pm0.14$ \\
19 &              B &         $1.0$ &           $0.8$ &              $20$ &     $0.43\pm0.14$ \\
\bottomrule
\end{tabular}
\end{adjustbox}

\caption{20 runs of segmentation hyperparameter tuning with \textbf{\textit{LOW}} granularity. Best and worst runs are highlighted in green and red. Metric evaluated on 250 test samples(mean/std).}
\label{tab:hyp_seg_low}

\end{table}

\begin{table}[!htb]
\centering
\begin{adjustbox}{height=0.25\linewidth}
\begin{tabular}{|c|c|c|c|c?c|c|}
\toprule
\multicolumn{5}{|c?}{\textbf{Hyperparameters}} & \multicolumn{1}{c|}{\textbf{Metrics}} \\

\midrule

\textbf{RUN} & \textbf{SP} & \textbf{BT} & \textbf{PT} & \textbf{SPS}& \textbf{DICE} \\
\midrule
0  &              P &         $0.6$ &           $0.8$ &               $5$ &      $0.2\pm0.07$  \\
1  &              P &         $0.8$ &           $1.0$ &              $10$ &     $0.21\pm0.07$  \\
2  &              P &         $1.0$ &           $0.6$ &               $5$ &      $0.2\pm0.08$  \\
3  &              P &         $0.8$ &           $0.8$ &              $10$ &     $0.21\pm0.08$  \\
4  &             PB &         $0.8$ &           $0.8$ &              $10$ &     $0.28\pm0.09$  \\
5  &              B &         $1.0$ &           $0.6$ &              $20$ &     $0.22\pm0.09$  \\
6  &             PB &         $0.8$ &           $0.8$ &              $10$ &     $0.27\pm0.09$  \\\rowcolor{red!20}
7  &              P &         $0.6$ &           $0.6$ &               $5$ &     $0.19\pm0.08$  \\
8  &             PB &         $0.8$ &           $1.0$ &              $10$ &     $0.27\pm0.09$  \\
9  &             PB &         $0.8$ &           $0.6$ &              $15$ &      $0.27\pm0.1$  \\
10 &              B &         $1.0$ &           $1.0$ &               $5$ &     $0.23\pm0.07$  \\
11 &              B &         $0.6$ &           $0.8$ &               $5$ &     $0.23\pm0.09$  \\
12 &              P &         $0.6$ &           $0.6$ &              $20$ &     $0.24\pm0.08$  \\
13 &             PB &         $0.6$ &           $0.6$ &               $5$ &     $0.25\pm0.11$  \\
14 &              P &         $1.0$ &           $0.6$ &              $10$ &     $0.21\pm0.08$  \\
15 &             PB &         $0.6$ &           $0.8$ &              $20$ &      $0.27\pm0.1$  \\\rowcolor{green!20}
16 &             PB &         $1.0$ &           $0.8$ &              $10$ &     $0.27\pm0.09$  \\
17 &              B &         $1.0$ &           $1.0$ &              $10$ &     $0.24\pm0.08$  \\
18 &             PB &         $0.8$ &           $0.6$ &              $20$ &      $0.26\pm0.1$  \\
19 &              P &         $0.6$ &           $0.8$ &              $20$ &     $0.24\pm0.08$  \\
\bottomrule
\end{tabular}
\end{adjustbox}

\caption{20 runs of segmentation hyperparameter tuning with \textbf{\textit{MEDIUM}} granularity. Best and worst runs are highlighted in green and red. Metric evaluated on 250 test samples(mean/std).}
\label{tab:hyp_seg_medium}

\end{table}

\begin{table}[!htb]
\centering
\begin{adjustbox}{height=0.25\linewidth}
\begin{tabular}{|c|c|c|c|c?c|c|}
\toprule
\multicolumn{5}{|c?}{\textbf{Hyperparameters}} & \multicolumn{1}{c|}{\textbf{Metrics}} \\

\midrule
\textbf{RUN} & \textbf{SP} & \textbf{BT} & \textbf{PT} & \textbf{SPS}& \textbf{DICE} \\
\midrule
0  &              B &         $0.6$ &           $0.6$ &              $15$ &     $0.18\pm0.06$  \\
1  &             PB &         $1.0$ &           $0.6$ &              $10$ &      $0.2\pm0.07$  \\
2  &              P &         $0.6$ &           $0.8$ &               $5$ &     $0.16\pm0.05$  \\
3  &              B &         $1.0$ &           $0.6$ &              $20$ &     $0.17\pm0.07$  \\
4  &              B &         $0.6$ &           $0.8$ &              $10$ &     $0.18\pm0.06$  \\
5  &              P &         $1.0$ &           $0.8$ &              $15$ &     $0.17\pm0.06$  \\
6  &             PB &         $0.6$ &           $0.6$ &              $15$ &     $0.21\pm0.07$  \\
7  &              P &         $0.8$ &           $1.0$ &              $20$ &     $0.18\pm0.06$  \\
8  &              B &         $0.8$ &           $1.0$ &              $20$ &     $0.18\pm0.06$  \\
9  &             PB &         $1.0$ &           $0.6$ &              $20$ &     $0.19\pm0.07$  \\
10 &              B &         $0.8$ &           $1.0$ &              $15$ &     $0.18\pm0.06$  \\
11 &             PB &         $0.6$ &           $0.8$ &               $5$ &      $0.2\pm0.06$  \\
12 &             PB &         $0.6$ &           $0.8$ &              $15$ &     $0.21\pm0.07$  \\
13 &              B &         $1.0$ &           $1.0$ &               $5$ &     $0.17\pm0.06$  \\
14 &             PB &         $1.0$ &           $1.0$ &               $5$ &     $0.19\pm0.07$  \\
15 &              P &         $0.6$ &           $0.8$ &              $20$ &     $0.18\pm0.07$  \\
16 &              P &         $0.8$ &           $0.8$ &              $20$ &     $0.18\pm0.07$  \\\rowcolor{red!20}
17 &              B &         $1.0$ &           $0.6$ &               $5$ &     $0.17\pm0.06$  \\
18 &              P &         $0.8$ &           $0.8$ &              $10$ &     $0.16\pm0.06$  \\\rowcolor{green!20}
19 &             PB &         $0.8$ &           $0.8$ &              $15$ &     $0.21\pm0.07$  \\
\bottomrule
\end{tabular}
\end{adjustbox}

\caption{20 runs of segmentation hyperparameter tuning with \textbf{\textit{HIGH}} granularity. Best and worst runs are highlighted in green and red. Metric evaluated on 250 test samples(mean/std).}
\label{tab:hyp_seg_high}

\end{table}

\clearpage

\clearpage
\section{Samples: \textit{Truck}}

\begin{figure*}[!htb]
    \centering
    \includegraphics[width=\textwidth]{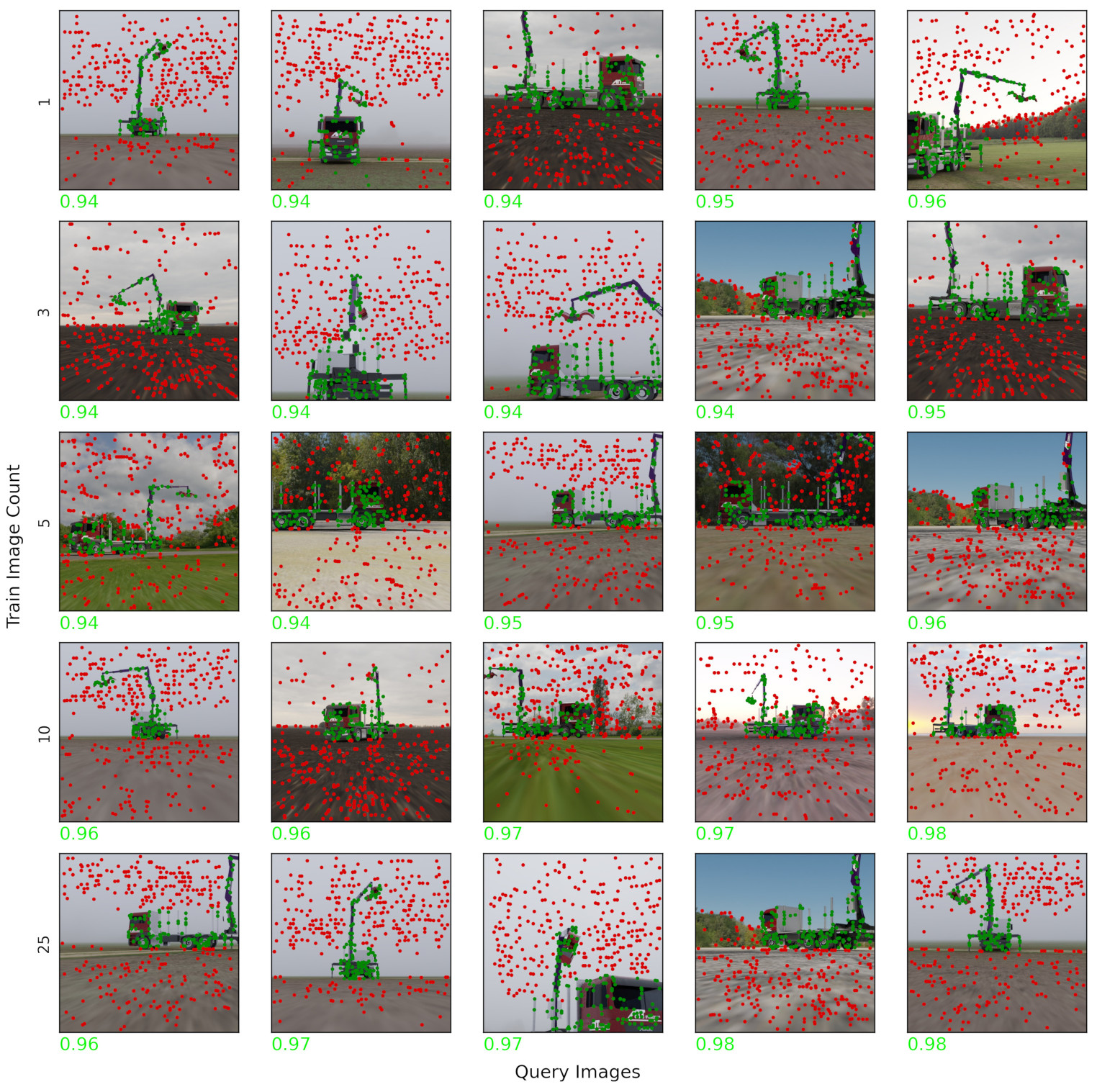}
    \caption{\textbf{\textit{Truck}} granularity graph node classification results for different train sample sizes. Each row shows the best five classification results for the given train sample size measured with \textit{F1} score.}
    \label{fig:ClassResultTruckSup} 
\end{figure*}

\begin{figure*}[!htb]
    \centering
    \includegraphics[width=\textwidth]{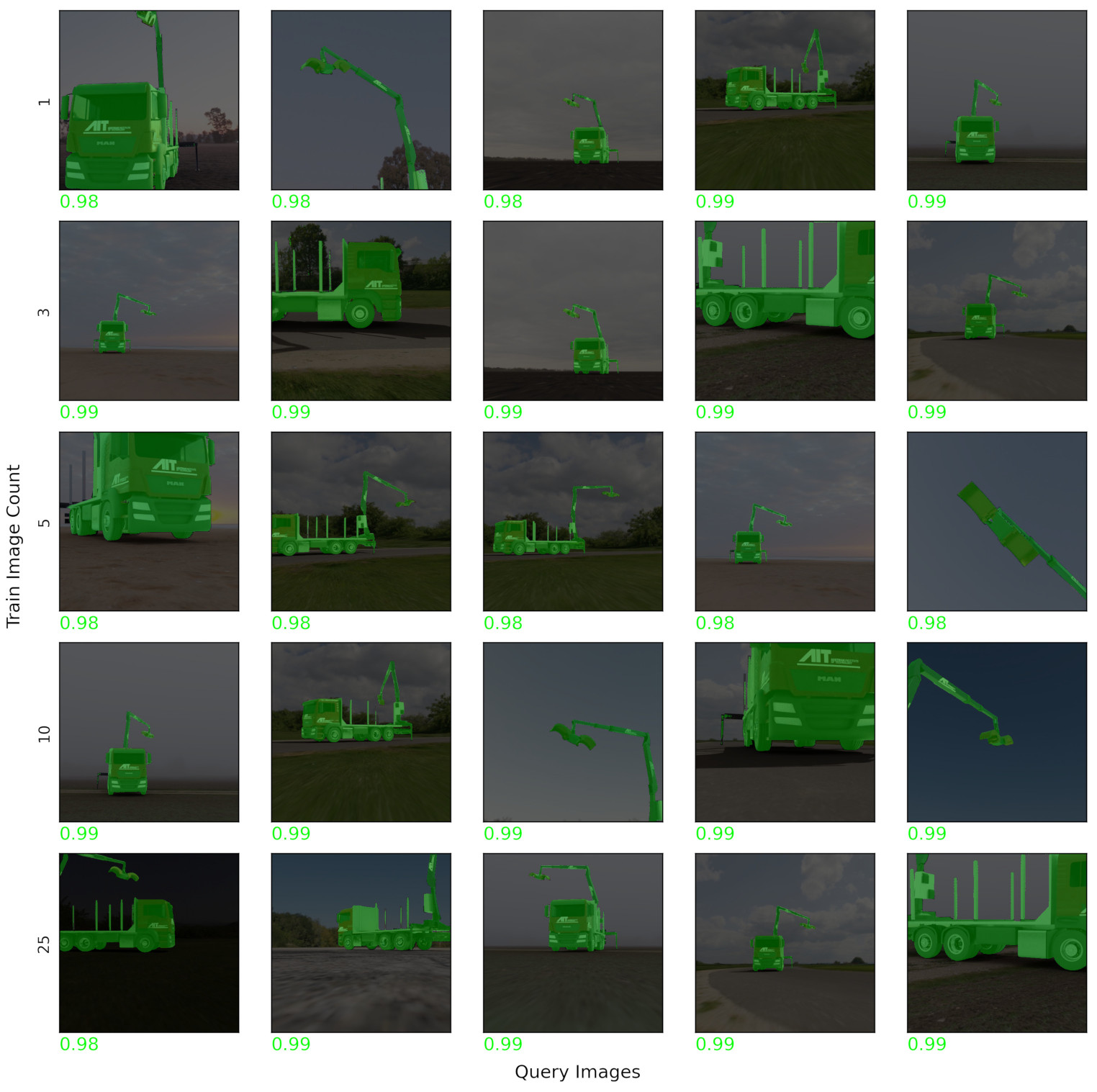}
    \caption{\textbf{\textit{Truck}} granularity semantic segmentation results for different train sample sizes. Each row shows the best five segmentation results for the given train sample size measured with \textit{dice} score.}
    \label{fig:SegResultTruckSup} 
\end{figure*}
\clearpage

\section{Samples: \textit{Truck Crane}}
\begin{figure*}[!htb]
   \centering
    \includegraphics[width=\textwidth]{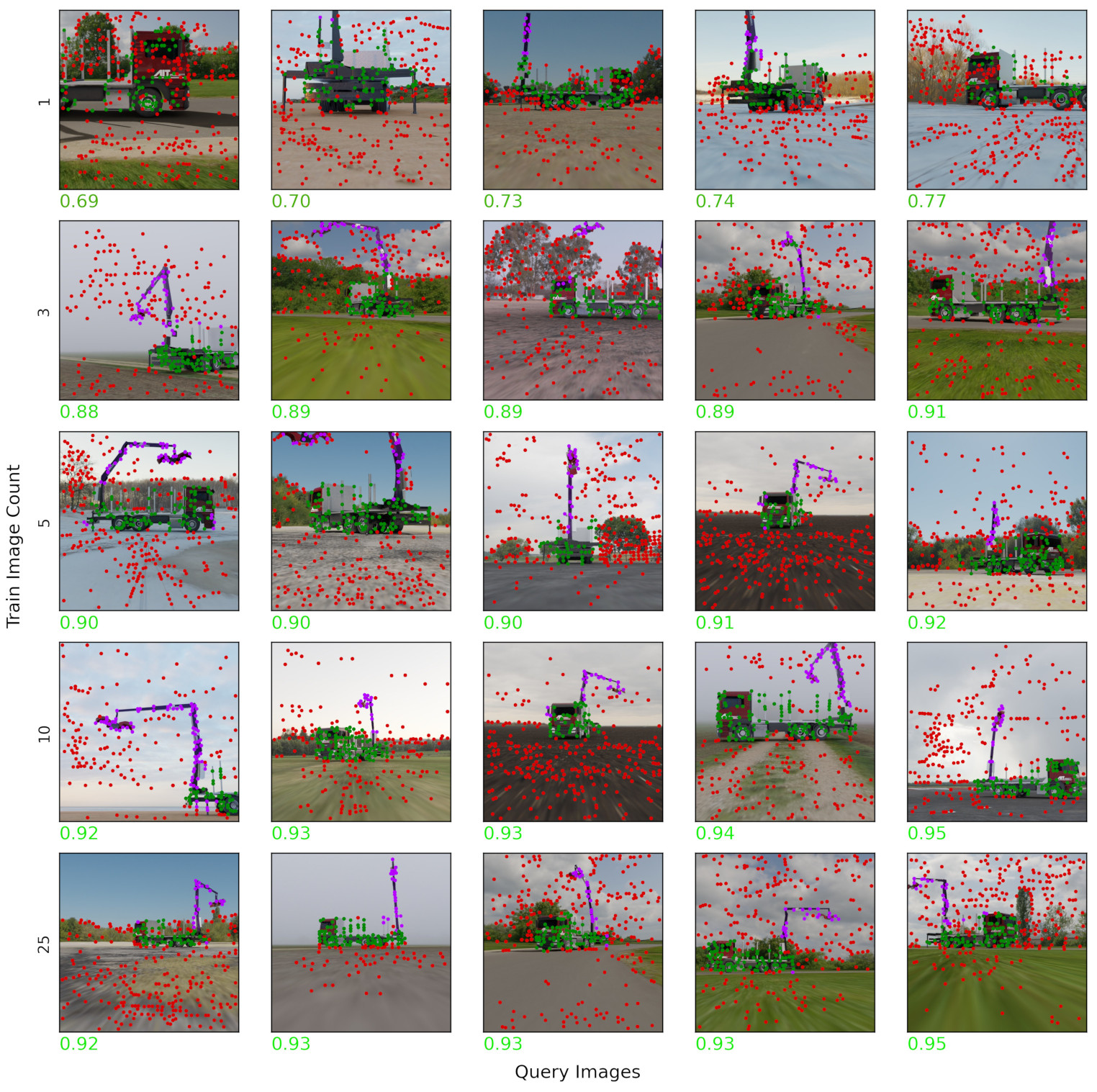}
    \caption{\textbf{\textit{Truck Crane}} granularity graph node classification results for different train sample sizes. Each row shows the best five classification results for the given train sample size measured with \textit{F1} score.}
    \label{fig:ClassResultTruckCraneSup}  
\end{figure*}
\begin{figure*}[!htb]
   \centering
    \includegraphics[width=\textwidth]{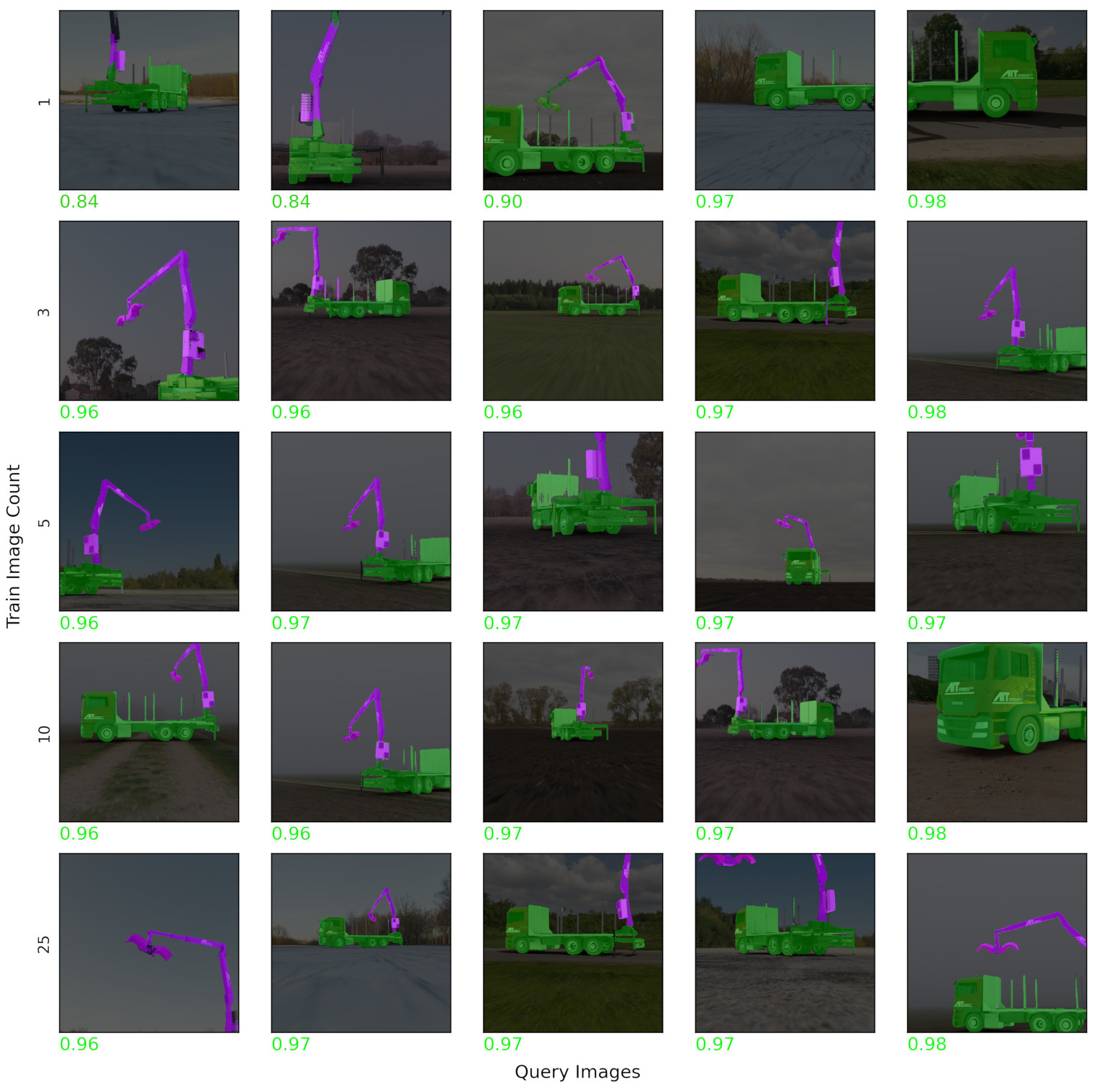}
    \caption{\textbf{\textit{Truck Crane}} granularity semantic segmentation results for different train sample sizes. Each row shows the best five segmentation results for the given train sample size measured with \textit{dice} score.}
    \label{fig:SegResultTruckCraneSup}  
\end{figure*}
\clearpage

\section{Samples: \textit{Low}}
\begin{figure*}[!htb]
    \centering
    \includegraphics[width=\textwidth]{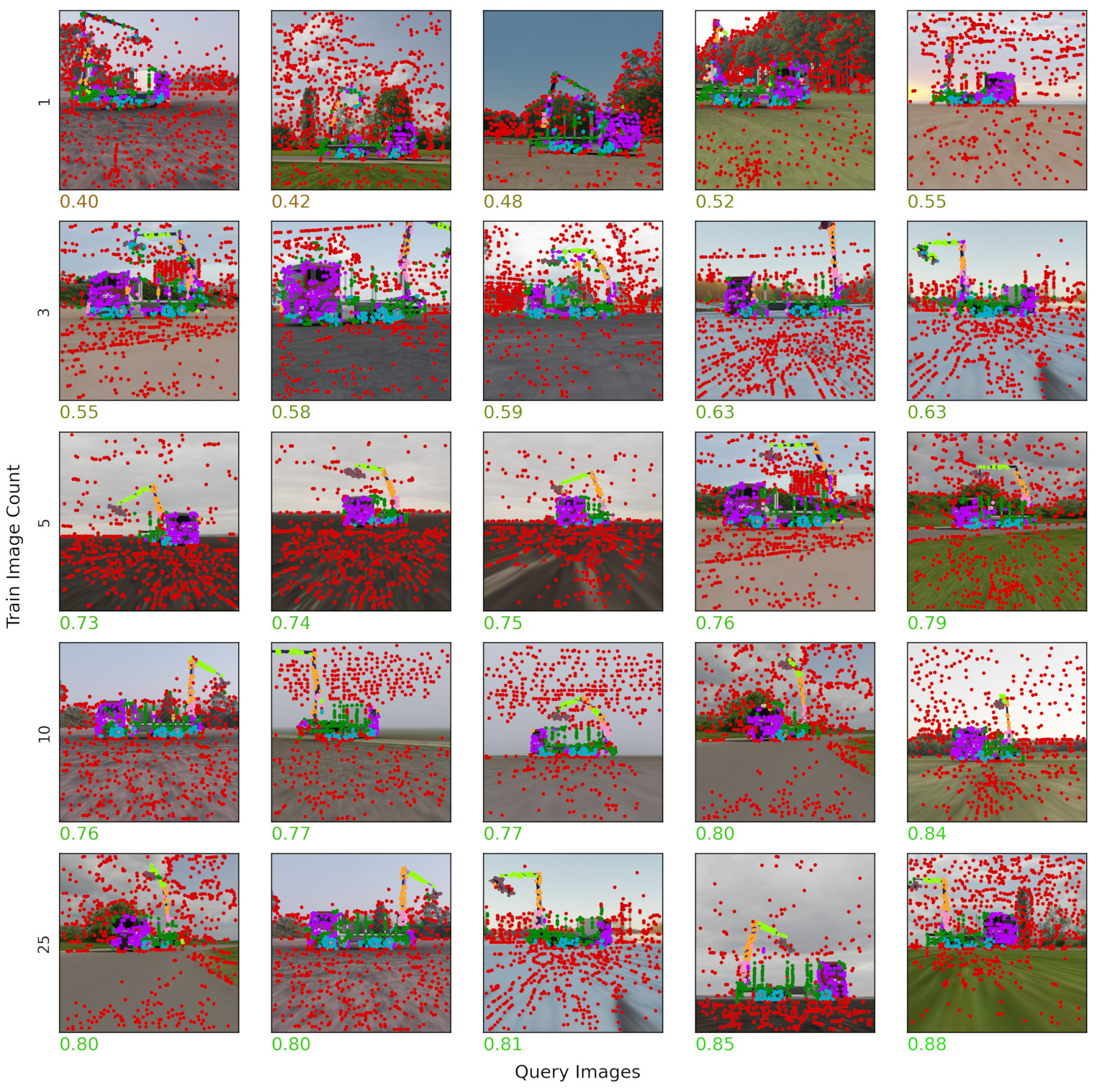}
    \caption{\textbf{\textit{Low}} granularity graph node classification results for different train sample sizes. Each row shows the best five classification results for the given train sample size measured with \textit{F1} score.}
    \label{fig:ClassResultLowSup}  
\end{figure*}
\begin{figure*}[!htb]
    \centering
    \includegraphics[width=\textwidth]{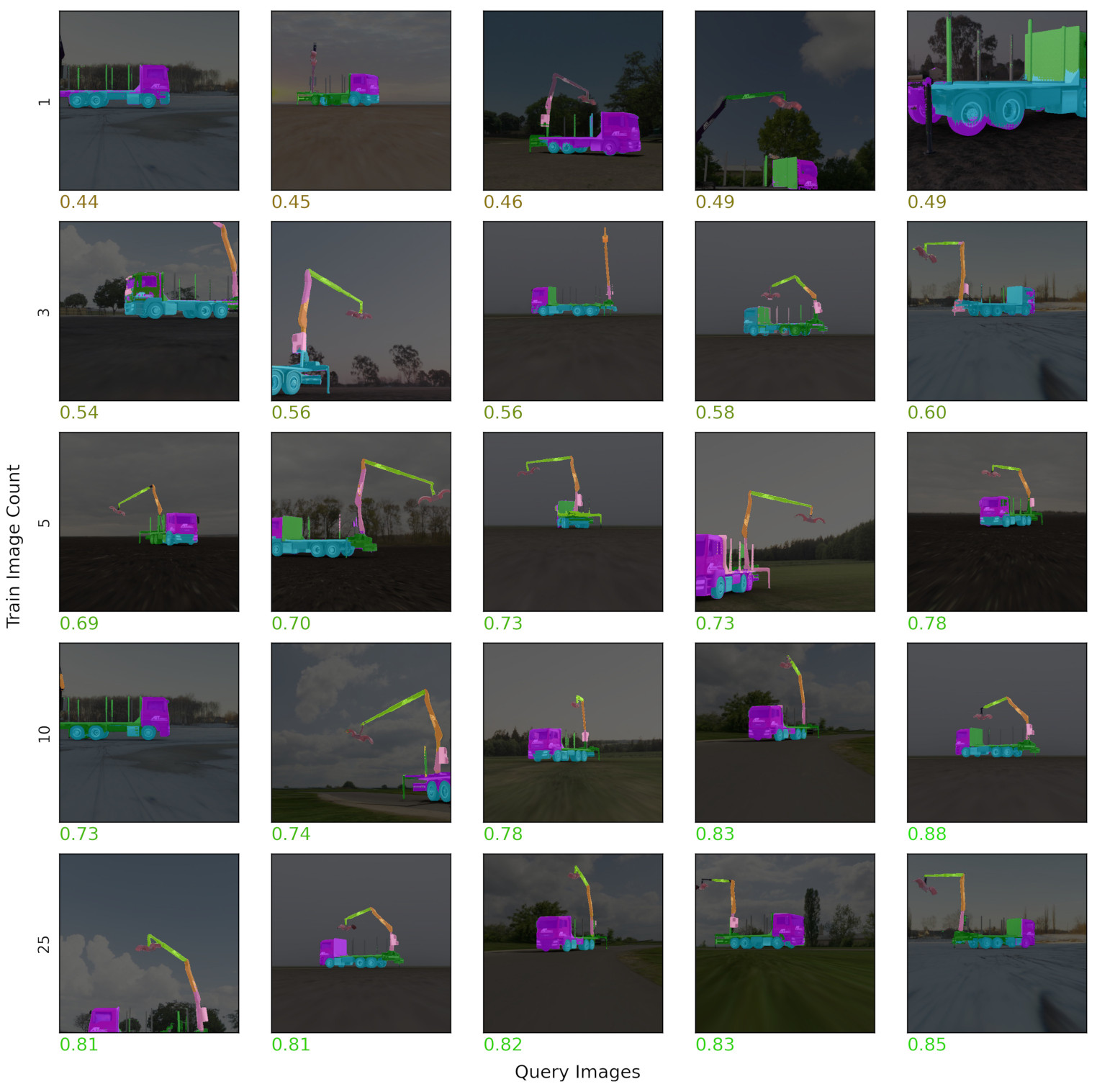}
    \caption{\textbf{\textit{Low}} granularity semantic segmentation results for different train sample sizes. Each row shows the best five segmentation results for the given train sample size measured with \textit{dice} score.}
    \label{fig:SegResultLowSup}  
\end{figure*}
\clearpage

\section{Samples: \textit{Medium}}
\begin{figure*}[!htb]
    \centering
    \includegraphics[width=\textwidth]{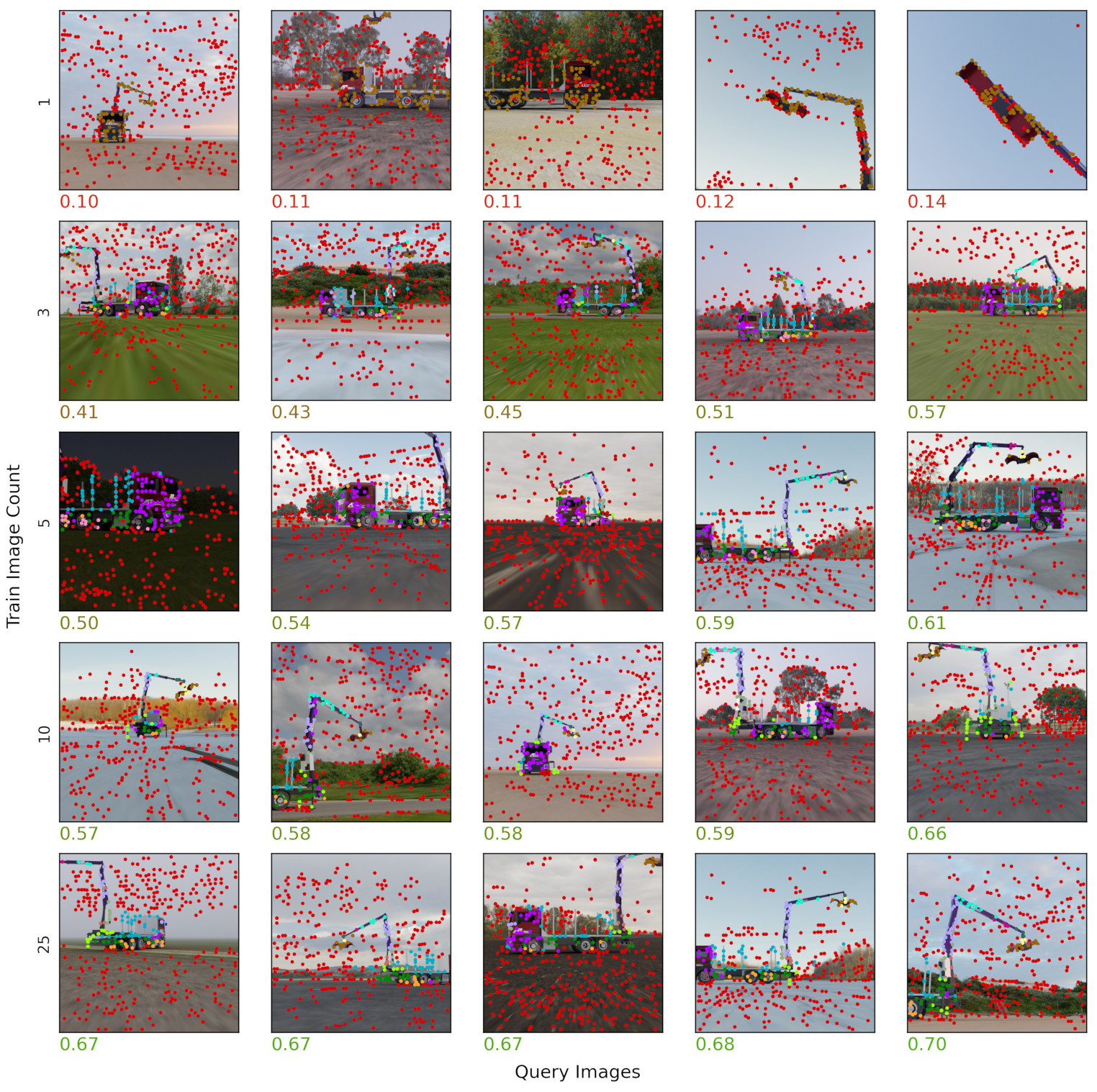}
    \caption{\textbf{\textit{Medium}} granularity graph node classification results for different train sample sizes. Each row shows the best five classification results for the given train sample size measured with \textit{F1} score.}
    \label{fig:ClassResultMediumSup}  
\end{figure*}
\begin{figure*}[!htb]
    \centering
    \includegraphics[width=\textwidth]{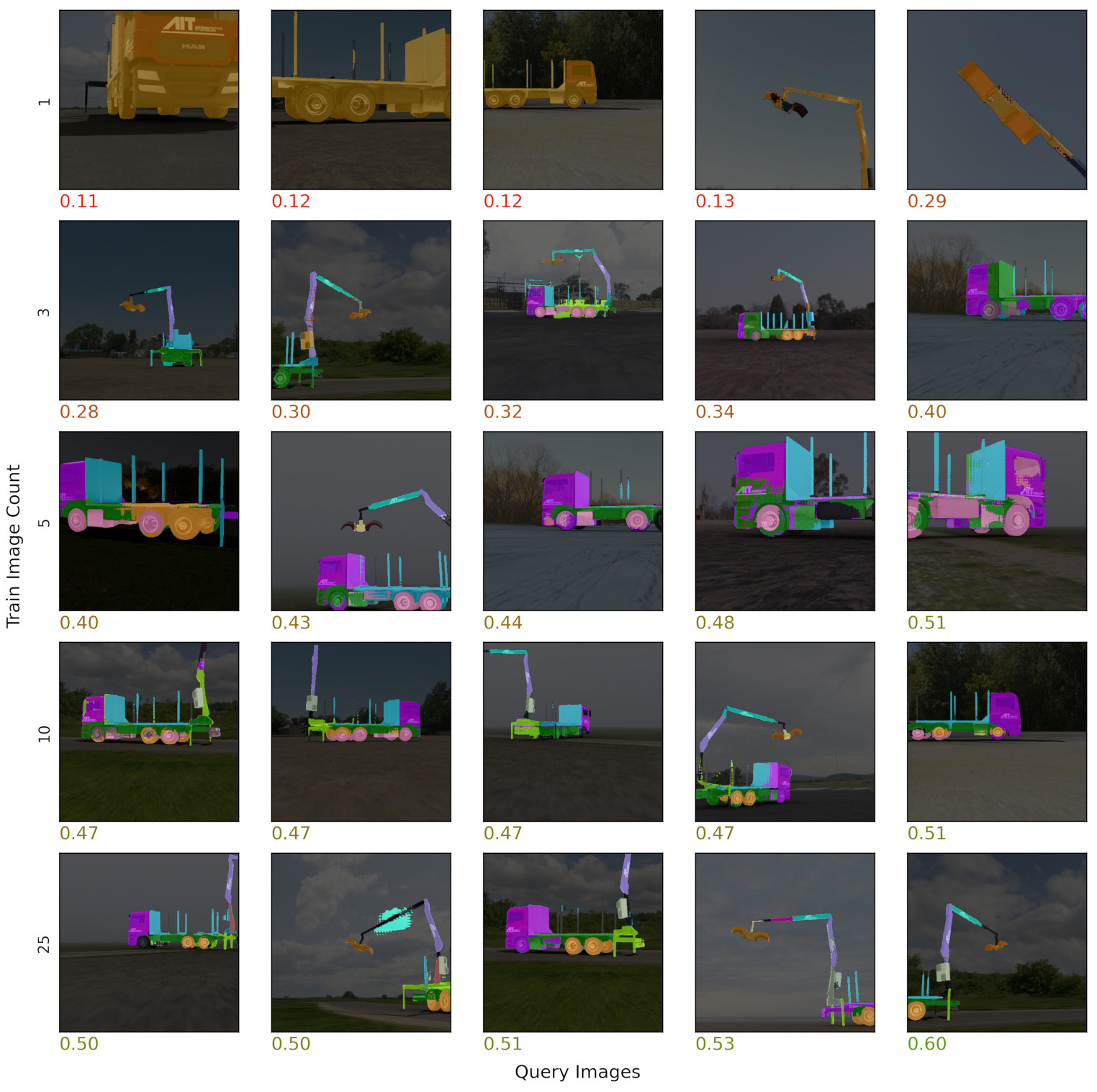}
    \caption{\textbf{\textit{Medium}} granularity semantic segmentation results for different train sample sizes. Each row shows the best five segmentation results for the given train sample size measured with \textit{dice} score.}
    \label{fig:SegResultMediumSup}  
\end{figure*}
\clearpage

\section{Samples: \textit{High}}
\begin{figure*}[!htb]
    \centering
    \includegraphics[width=\textwidth]{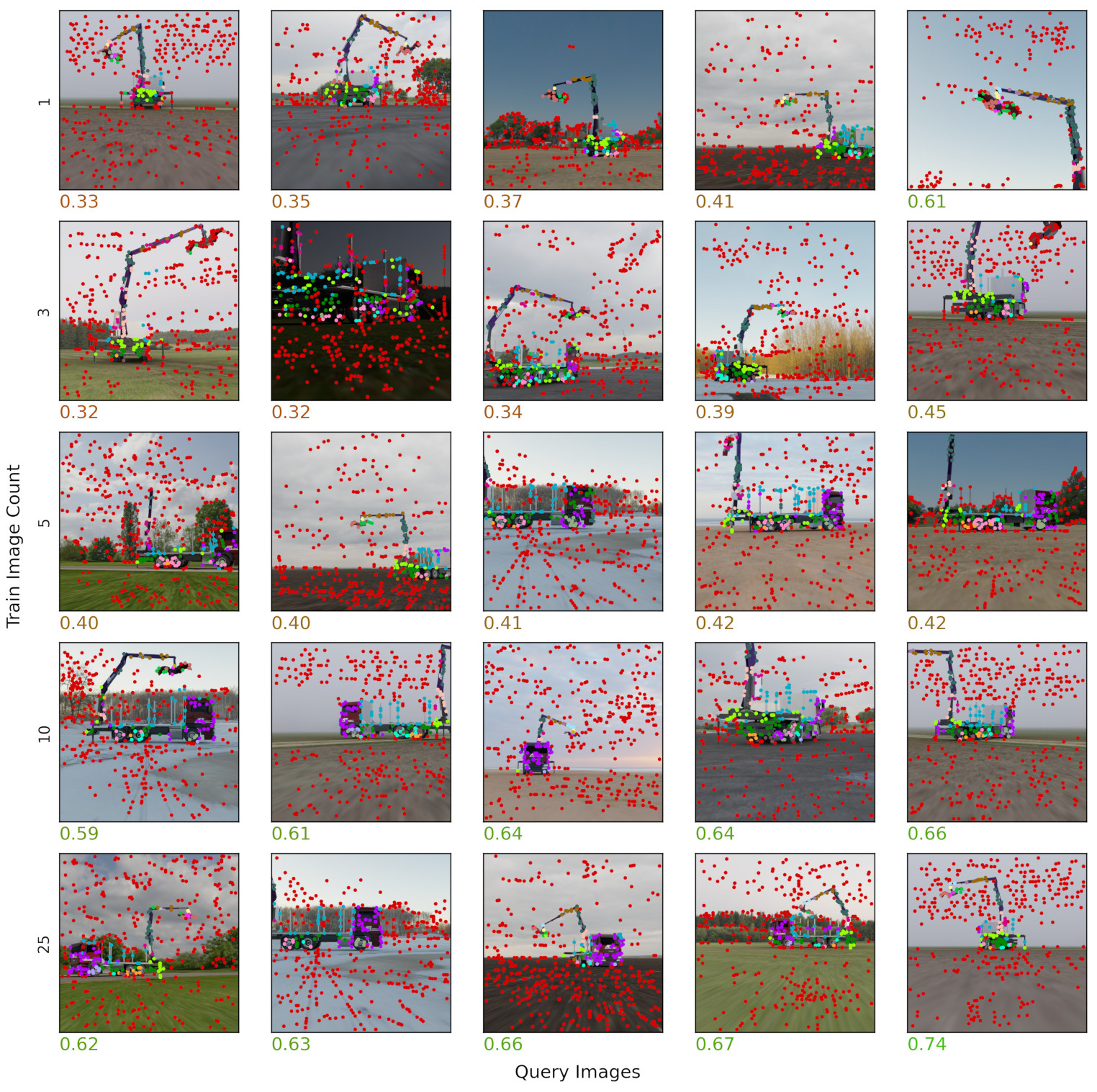}
    \caption{\textbf{\textit{High}} granularity graph node classification results for different train sample sizes. Each row shows the best five classification results for the given train sample size measured with \textit{F1} score.}
    \label{fig:ClassResultHighSup}  
\end{figure*}
\begin{figure*}[!htb]
    \centering
    \includegraphics[width=\textwidth]{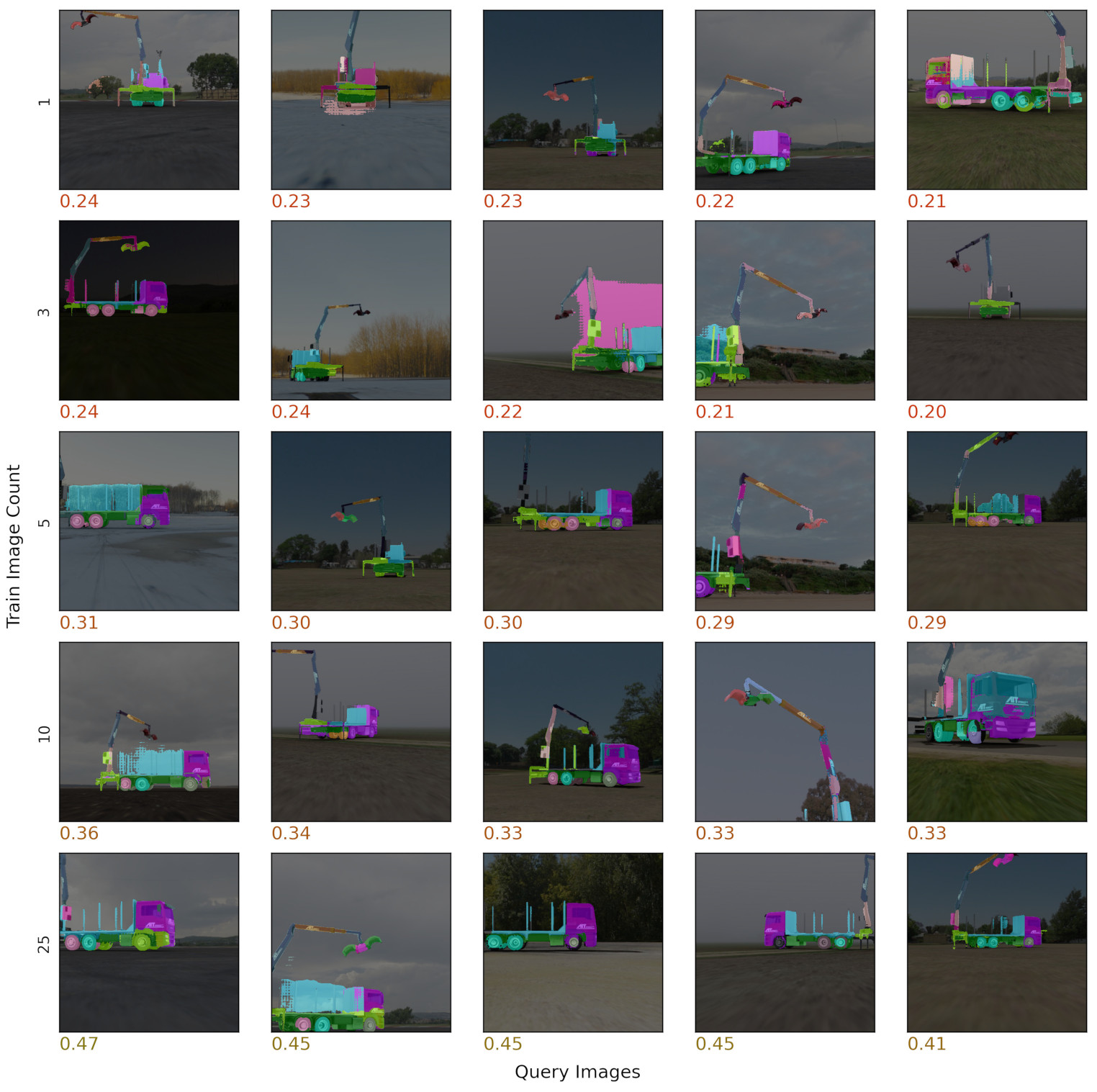}
    \caption{\textbf{\textit{High}} granularity semantic segmentation results for different train sample sizes. Each row shows the best five segmentation results for the given train sample size measured with \textit{dice} score.}
    \label{fig:SegResultHighSup}  
\end{figure*}

\clearpage

\section{Mask R-CNN comparison} \label{apx:maskrcnn_comparison}

\begin{figure}[!htb]
    \centering
    \includegraphics[width=1\linewidth]{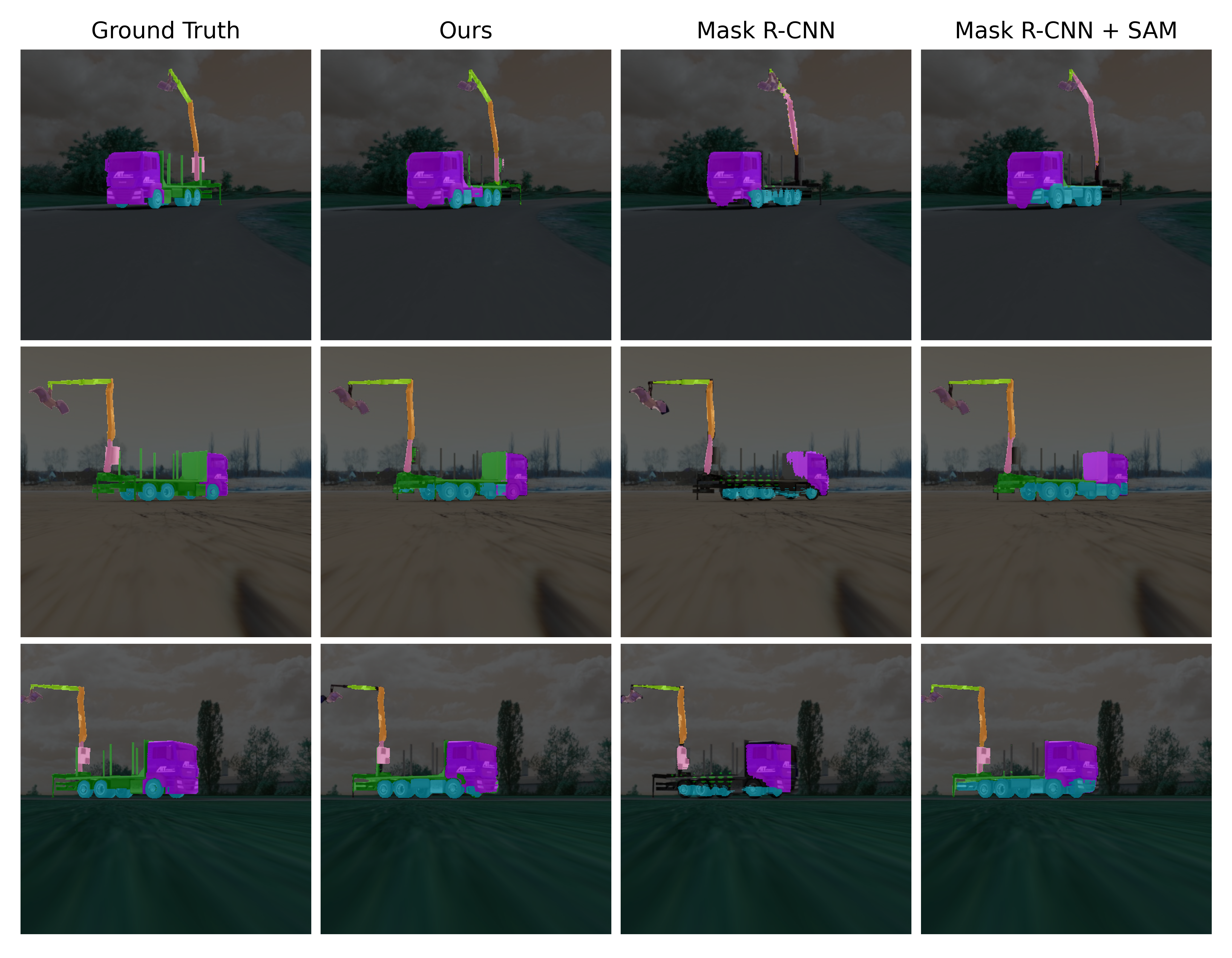}
    \caption{Our method, trained on just 10 samples, demonstrates competitive performance when compared to fine-tuning pre-trained (COCO) models like Mask R-CNN and Mask R-CNN + SAM (which uses estimated bounding boxes as SAM prompts). While Mask R-CNN was trained with 15 training and 10 validation samples, our method achieved superior results with fewer training data. This highlights the robustness of our approach, especially in low-data scenarios, where conventional methods like Mask R-CNN typically require larger datasets to perform optimally.}
    \label{fig:enter-labelSup}
\end{figure}

\clearpage


\end{document}